\documentclass{article}

\usepackage{tech2025}

\usepackage{microtype}
\usepackage{graphicx}
\usepackage{subfigure}
\usepackage{booktabs} 

\usepackage{hyperref}

\usepackage{amsmath}
\usepackage{amssymb}
\usepackage{mathtools}
\usepackage{amsthm}
\usepackage{mathrsfs}
\usepackage{xspace}
\usepackage{academicons}

\usepackage[capitalize,noabbrev]{cleveref}

\theoremstyle{plain}

\theoremstyle{definition}

\theoremstyle{remark}

\definecolor{bigaired}{RGB}{156, 0, 0}
\definecolor{uclablue}{RGB}{39, 116, 174}
\definecolor{thupurple}{RGB}{102, 8, 116}
\definecolor{pkured}{RGB}{139, 0, 18}
\definecolor{panton}{RGB}{217, 51, 121}

\definecolor{darkred}{RGB}{200, 0, 0}
\definecolor{darkblue}{RGB}{0, 0, 200}
\definecolor{blue}{RGB}{0, 0, 250}

\definecolor{light}{RGB}{225, 250, 250}
\definecolor{lightgray}{RGB}{0.9, 0.9, 0.9}
\definecolor{lightred}{RGB}{250, 200, 200}
\definecolor{lightblue}{RGB}{210, 220, 250}
\definecolor{lightpurple}{RGB}{218,210,255}

\definecolor{doderblue}{RGB}{30, 144, 255}
\definecolor{select}{RGB}{222, 235, 247}
\definecolor{unselect}{RGB}{247, 207, 206}

\definecolor{myLinkColor}{HTML}{7D5BA6}     
\definecolor{myCiteColor}{rgb}{0,0.08,0.45} 
\definecolor{myURLColor}{HTML}{5B7DB1}      
\hypersetup{
  colorlinks=true,
  linkcolor=myLinkColor,
  citecolor=myCiteColor,
  urlcolor=myURLColor
}

\usepackage[textsize=tiny]{todonotes}

\usepackage[most]{tcolorbox}
\usepackage{fontawesome5}  
\usepackage{listings}      
\usepackage[misc]{ifsym}
\usepackage{wrapfig}    
\usepackage[T1]{fontenc}

\usepackage{tcolorbox}
\usepackage{relsize}

\usepackage{adjustbox}
\usepackage{fancyhdr}
\usepackage{lipsum}
\usepackage{newtxtext}
\usepackage{wrapfig}
\usepackage{enumitem}
\definecolor{azblue}{RGB}{27,117,187}      

\usepackage[noend]{algpseudocode}   

\definecolor{bestcol}{RGB}{  0,102,204} 
\definecolor{goodcol}{RGB}{ 34,139, 34} 
\definecolor{deltaBg}{RGB}{220,230,255} 

\usepackage[T1]{fontenc}
\usepackage[utf8]{inputenc}
\usepackage{lmodern}  

\usepackage{tikz}
\usetikzlibrary{tikzmark,decorations.pathreplacing,calc}
\usepackage{stackengine}
\stackMath
\definecolor{lightgreen}{RGB}{0,150,0}  

\usepackage{titletoc}

\newtheoremstyle{rqstyle}%
  {\topsep}            
  {\topsep}            
  {}                   
  {}                   
  {\bfseries}    
  {:}                  
  {.5em}               
  {}                   

\theoremstyle{rqstyle}

\crefname{researchquestion}{Research Question}{Research Questions}

\definecolor{propose}{HTML}{EF8E8D}
\definecolor{solve}{HTML}{5755A3}

\definecolor{humanred}{RGB}{180, 50, 50}
\definecolor{envgreen}{RGB}{50, 140, 80}
\definecolor{lightblue}{RGB}{224,234,245}
\definecolor{paleviolet}{HTML}{E1EEFC}
\definecolor{lightgrey}{RGB}{247, 247, 247}
\newenvironment{leapabstract}{
  \begin{tcolorbox}[
    colback=lightblue,
    colframe=white,
    boxrule=0pt,
    arc=10pt,
    left=16pt,
    right=16pt,
    top=12pt,
    bottom=12pt,
    width=\textwidth,
    enlarge left by=0mm,
    before skip=10pt,
    after skip=10pt
  ]
  \normalsize
}{
  \end{tcolorbox}
}

\makeatletter
\DeclareRobustCommand\onedot{\futurelet\@let@token\@onedot}
\def\@onedot{\ifx\@let@token.\else.\null\fi\xspace}

 \def\vs{\emph{vs}\onedot}

\makeatother

\begin{document}


\makeatletter
\def\icmldate#1{\gdef\@icmldate{#1}}
\icmldate{}
\makeatother


\makeatother

\vspace*{-3em}
\icmltitle{ReasonMed: A 370K Multi-Agent Generated Dataset for Advancing Medical Reasoning}

\begin{icmlauthorlist}
\mbox{Yu Sun$^{\,1,2,\dagger}$}, 
\mbox{Xingyu Qian$^{\,1,3,4,5,\dagger}$}, 
\mbox{Weiwen Xu$^{\,1}$},
\mbox{Hao Zhang$^{\,1}$},
\mbox{Chenghao Xiao$^{\,1}$},  
\mbox{Long Li$^{\,1}$},
\mbox{Deli Zhao$^{\,1}$},
\mbox{Wenbing Huang$^{\,3,4,5}$}, 
\mbox{Tingyang Xu$^{\,1,\,6,\ddagger}$} 
\mbox{Qifeng Bai$^{\,2,\ddagger}$}, 
\mbox{Yu Rong$^{\,1,\,6,\ddagger}$}

\end{icmlauthorlist}

$^{1\,}$Alibaba DAMO Academy \quad
$^{2\,}$School of Basic Medical Sciences, Lanzhou University \quad
$^{3\,}$Gaoling School of Artificial Intelligence, Renmin University of China\quad
$^{4\,}$Beijing Key Laboratory of Research on Large Models and Intelligent Governance\quad
$^{5\,}$Engineering Research Center of Next-Generation Intelligent Search and Recommendation, MOE\quad
$^{6\,}$Hupan Lab\\
$^{\dagger}$Equal Contribution \quad
$^{\ddagger}$Corresponding Author

{\smaller \texttt{yusunaiwork@gmail.com}, ~\texttt{hzhang26@outlook.com}, ~\texttt{baiqf@lzu.edu.cn}, ~\texttt{xuty\_007@hotmail.com}}

\icmlcorrespondingauthor{Qifeng Bai}{baiqf@lzu.edu.cn}
\icmlcorrespondingauthor{Tingyang Xu}{xuty\_007@hotmail.com}
\vspace*{0.7em}

\begin{leapabstract}
Reasoning-based large language models have excelled in mathematics and programming, yet their potential in knowledge-intensive medical question answering remains underexplored and insufficiently validated in clinical contexts.
To bridge this gap, we introduce \textbf{ReasonMed}, the largest medical reasoning dataset to date, comprising 370k high-quality examples distilled from 1.75 million initial reasoning paths generated by complementary LLMs and curated through a cost-efficient easy-medium-difficult (EMD) pipeline.
ReasonMed is built through a multi-agent generation, verification, and refinement process, in which an \textit{Error Refiner} improves reasoning paths by correcting error-prone steps identified by a verifier.
Using ReasonMed, we investigate effective strategies for training medical reasoning models and find that integrating detailed CoT reasoning with concise answer summaries yields the most robust fine-tuning results.
Models trained on ReasonMed set a new benchmark: ReasonMed-7B surpasses the prior best sub-10B models by 4.17\% and even exceeds LLaMA3.1-70B on PubMedQA by 4.60\%. When scaled to ReasonMed-14B, it remains highly competitive, underscoring consistent scaling potential.

\vspace{0.9em}

\makebox[\textwidth]{
  \hspace{-0.3em}
  \raisebox{-0.9ex}{%
      \begin{tabular}{c}
        \href{https://github.com/YuSun-Work/ReasonMed}{\includegraphics[height=1.75em]{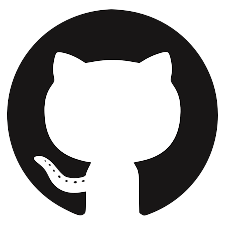}} \\
        \textbf{\scriptsize \href{https://github.com/YuSun-Work/ReasonMed}{Code}}
      \end{tabular}
  } \\
  \hspace{-0.5em}
  \raisebox{-0.9ex}{%
    
      \begin{tabular}{c}
        \href{https://github.com/YuSun-Work/ReasonMed}{\includegraphics[height=1.75em]{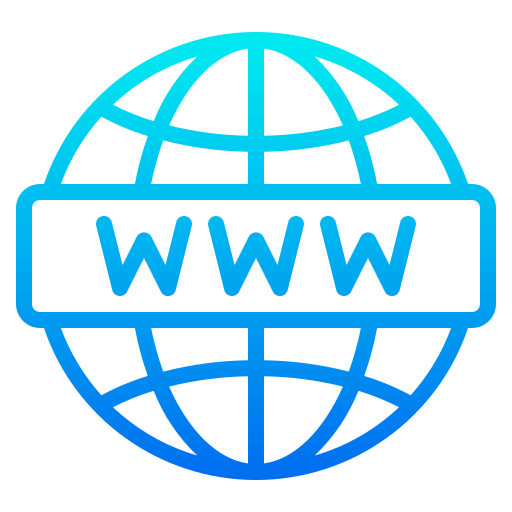}} \\
        \textbf{\scriptsize \href{https://github.com/YuSun-Work/ReasonMed}{Project Page}}
      \end{tabular}
  } \\
  \hspace{-0.5em}
  \raisebox{-0.9ex}{%
      \begin{tabular}{c}
        \href{https://huggingface.co/datasets/YuSun-AI/ReasonMed}{\includegraphics[height=1.75em]{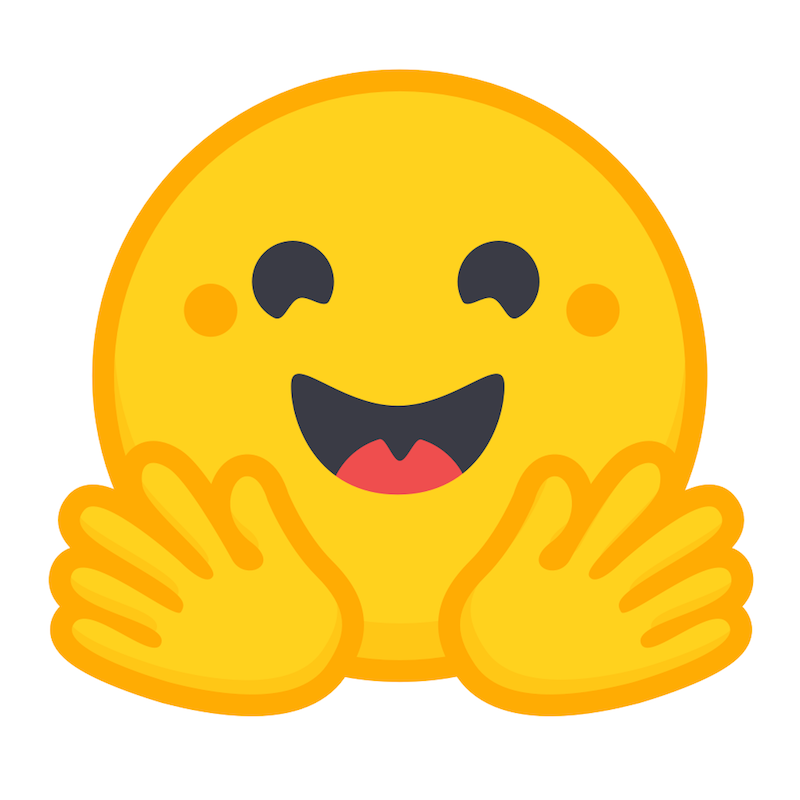}} \\
        \textbf{\scriptsize \href{https://huggingface.co/datasets/YuSun-AI/ReasonMed}{Dataset}}
      \end{tabular}
  }
}
\vskip -2cm
\end{leapabstract}

\section{Introduction}
\label{sec:introduction}

Recent reasoning-oriented large language models (LLMs; \citet{deepseekai2025deepseekr1incentivizingreasoningcapability,qwq32b}) have garnered significant attention due to their remarkable capabilities in logical reasoning~\cite{liu2025logicalreasoninglargelanguage,chen2025finereason}, mathematics~\cite{ahn2024largelanguagemodelsmathematical}, query optimization~\cite{tan2025can,huanshuo2025ctrla}, and programming~\cite{openai2025competitiveprogramminglargereasoning} tasks.

Despite their effectiveness, LLMs encounter distinct challenges in the medical domain due to its knowledge-intensive nature, which requires extensive, high-quality, and accurately curated data. Existing medical reasoning datasets, such as medical-o1-reasoning-SFT and Medical-R1-Distill-Data~\cite{chen2024huatuogpto1medicalcomplexreasoning}, remain limited in scale and typically derived from a single teacher model, restricting knowledge coverage. Moreover, prior work lacks a systematic analysis of the trade-offs between resource-intensive, multi-step reasoning~\cite{wei2023chainofthoughtpromptingelicitsreasoning} and more compact, summary-based strategies. Whether the added computational cost of explicit reasoning outweighs its benefits over efficient summarization in medical QA systems remains an open question.

To tackle these challenges, we present ReasonMed, a large-scale medical reasoning dataset containing 370k rigorously verified examples--an order of magnitude larger than prior efforts~\cite{chen2024huatuogpto1medicalcomplexreasoning}. Curated from multiple advanced LLMs, ReasonMed captures diverse medical insights with broad depth and coverage. Each entry provides detailed multi-step CoT reasoning alongside a concise answer summary, facilitating analysis of effective reasoning strategies in the medical domain.

Dataset scale plays a crucial role in enhancing model performance. To this end, we adopt a large-scale, high-quality data generation paradigm using a multi-agent system (MAS). We first aggregate approximately 195k questions (excluding test splits) from four benchmarks: MedQA~\cite{jin2020disease}, MMLU~\cite{hendryckstest2021}, PubMedQA~\cite{jin2019pubmedqa}, and MedMCQA~\cite{pmlr-v174-pal22a}.
Our MAS integrates three strong LLMs, \emph{i.e.}, two general-purpose (Qwen2.5-72B~\cite{qwen2.5} and DeepSeek-R1-Distill-Llama-70B~\cite{deepseekai2025deepseekr1incentivizingreasoningcapability}) and one medical-specific (HuatuoGPT-o1-70B~\cite{chen2024huatuogpto1medicalcomplexreasoning}). By varying sampling hyperparameters (\emph{e.g.}, temperature, top-p) across agents, we generate  $\sim$1.75 million diverse, multi-step reasoning paths. 

This combination of scale and methodological rigor ensures higher data quality and, in turn, stronger clinical QA performance.

Beyond data scale, training efficacy is highly sensitive to data quality. Prior work~\cite{muennighoff2025s1simpletesttimescaling} shows that even $1,000$ well-curated examples can yield strong results.
To ensure precision in medical QA, we devise a rigorous quality control pipeline that validates each reasoning chain for \textit{answer correctness}, \textit{logical coherence}, and \textit{medical factuality}. Based on validation pass rates, questions are classified into three tiers: \textit{easy} ($\geq$5 correct paths), \textit{medium} (2$\sim$4 correct paths), and \textit{difficult} ($<$2 correct paths). For easy cases, the two top-ranked reasoning paths verified by a quality ranker are retained. For medium cases, where subtle errors are more common, an \textit{error refiner}, guided by verifier logs and powered by GPT-4o-mini, revises and expands the paths. For difficult cases, GPT-o1 is employed with a structured multi-step process to directly generate valid reasoning. This multi-stage refinement yields a polished dataset of 370k high-quality medical reasoning samples.

In addition to curating high-quality reasoning data, we examine how different training strategies affect model performance. Specifically, we compare fine-tuning methods, including traditional CoT, summary-based responses, and a hybrid CoT-summary approach. Using the \texttt{lm\_eval} framework~\cite{eval-harness}, we identify the most effective strategies for enhancing medical LLMs on complex questions. Results show that the hybrid method achieves the best accuracy, while summary-only responses provide competitive results with lower computational cost, underscoring the value of strategy selection based on application needs.

To ensure medical validity beyond LLM-as-a-judge, we conduct a pilot review of 100 randomly sampled cases by board-certified physicians. Their evaluations align with our automatic scores and surface concrete strengths, such as structured reasoning (87\%) and systematic distractor elimination (82\%), alongside actionable gaps, including insufficient external citations (68\%) and limited clinical specifics (35\%).
We further demonstrate favorable scaling: a 14B model fine-tuned on ReasonMed surpasses competitive 14B/32B baselines and approaches 70B performance across 9 medical benchmarks. We release complete training and inference details, along with a temperature sensitivity analysis indicating peak accuracy at $0.5\sim0.7$.

Our main contributions are fourfold:
\begin{itemize}
    \item We release the largest open-source medical reasoning dataset, containing 1.29M validated paths, distilled to 370k high-quality examples through targeted optimization.
    \item We design a multi-agent framework for generating, filtering, and refining reasoning paths. GPT-4o evaluations on random subsets of $1,000$ and $3,000$ entries confirm ReasonMed's superior quality over data generated by GPT-4o and DeepSeek-R1.
    \item We present the first controlled study in knowledge-intensive medical QA using an identical data source, isolating the contributions of multi-step reasoning versus dense knowledge injection, and systematically assessing their effects on accuracy and efficiency.
    \item We train the ReasonMed-7B on reasoning-augmented data, achieving 82.0\% on PubMedQA, surpassing LLaMA3.1-70B by +4.6\%. Scaling to 14B, ReasonMed-14B outperforms Qwen2.5-14B by +3.8\%, exceeds Qwen2.5-32B, and approaches LLaMA3.1-70B, demonstrating stable scaling potential.
\end{itemize}

\section{Related Work}
\label{sec:related}

\begin{figure*}[ht]
  \includegraphics[width=\textwidth]{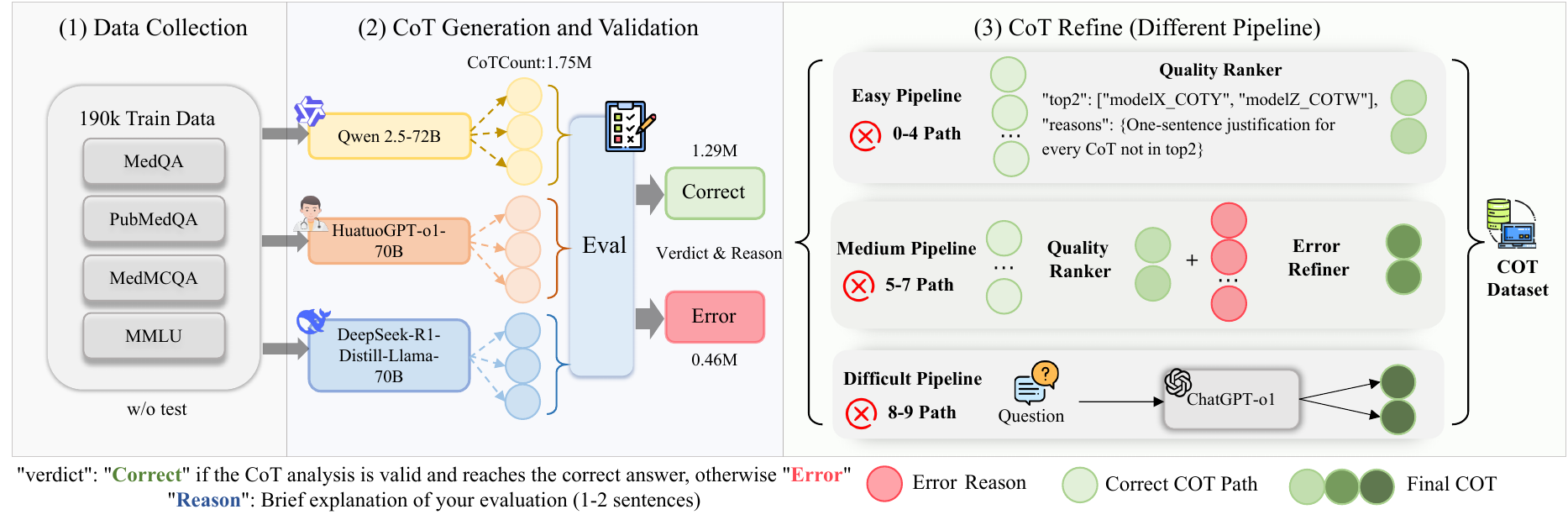}
  \caption{(1) show composition of the dataset. (2) present the Multi-Agent System for generating and validating Complex CoT. (3) outline strategy schemes (Easy/Medium/Difficult Pipeline) based on CoT validation counts. For 0-4 errors, select top two CoTs using the Quality Ranker. For 5-7 errors, optimize the top two CoTs with GPT-4o-mini, addressing identified weak points. For 8-9 errors, generate high-quality answers using GPT-o1.}
  \label{fig:Pipeline}
\end{figure*}

\paragraph{Multi-Agent-based Data Curation.}
Multi-agent frameworks have proven effective and robust for dataset generation and optimization across diverse domains. These systems assign specialized roles to agents that collaborate in a team-like manner~\cite{hong2023metagpt,li2025amaze,gu2025rapid,qi2024long2rag}. For instance, DialogueAgents~\cite{li2025dialogueagentshybridagentbasedspeech} employs scriptwriters, synthesizers, and critics to produce diverse, high-quality dialogue data, while AgentCoder~\cite{huang2024agentcodermultiagentbasedcodegeneration} leverages programmers, test designers, and executors to iteratively refine code datasets through agent-driven feedback. Similarly, BOLT~\cite{pang2025boltbootstraplongchainofthought} integrates multi-agent frameworks with LLMs to generate long-chain reasoning data, underscoring their utility in structured, reasoning-intensive tasks. Distinct from prior work, our framework targets medical reasoning datasets, combining domain-specific and general-purpose language models to generate, validate, and refine reasoning paths tailored for medical QA.

\paragraph{Medical Reasoning Dataset \& Model.}
Recent work demonstrates the effectiveness of chain-of-thought (CoT) prompting for medical QA~\cite{wei2022medicalCOT,liévin2023largelanguagemodelsreason}. Adaptive reasoning models, such as medical language agents, have been proposed to address complex clinical tasks~\cite{dutta2024adaptive}. Multi-agent frameworks further enhance reliability and interpretability by coordinating specialized medical reasoning agents~\cite{zuo2025mutliagentreasoning}. HuatuoGPT~\cite{chen2024huatuogpto1medicalcomplexreasoning} exemplifies the integration of rich medical knowledge with multi-step reasoning in large models.
In parallel, Lingshu introduces a unified multimodal foundation model that jointly handles text- and image-based clinical problems, achieving strong performance on multimodal QA and related benchmarks~\cite{xu2025lingshu}. While Lingshu emphasizes unified modeling and evaluation, our work targets the creation of high-quality CoT datasets for medical QA through rigorous verification and staged optimization. Existing datasets typically lack systematic verification and structured optimization tailored to medical reasoning. We address this gap with a multi-stage pipeline that evaluates, verifies, and refines reasoning paths, producing datasets with substantially improved fidelity and applicability.

\paragraph{LLM-as-a-Judge.}
Large language models are increasingly adopted as evaluators (LLM-as-a-Judge), offering scalable and consistent assessment across domains~\cite{gu2025surveyllmasajudge,zeng2024mrben}. In medical QA, they improve evaluation reliability and accuracy~\cite{krolik2024leveraginglargelanguagemodels,zhao2024artificial}. By iteratively reviewing reasoning steps, LLM evaluators guide models toward coherent and correct solutions~\cite{qin2024o1replicationjourneystrategic}. Frameworks such as QuRating~\cite{YiDaTang2024TheInnovation} highlight their utility in systematically selecting high-quality training data. Distinct from prior work, our method evaluates language-model-generated CoT reasoning for both correctness and factual fidelity, while explicitly identifying errors to support targeted optimization. We further introduce a score-based evaluator that quantifies reasoning improvements after refinement and measures overall dataset quality.

\section{Multi-Agent Reasoning Pipeline}

\subsection{Dataset Composition}
In this section, we present the composition of the dataset used for the Multi-Agent Reasoning Pipeline, along with an analysis of the dataset's structure and the benchmarks involved. The dataset consists of various medical question-answering datasets. Table \ref {tab:dataset-summary} shows a summary of the dataset composition:

\begin{table}[ht]
\centering
\caption{Summary of ReasonMed Question Count Composition.}
\label{tab:dataset-summary}
\begin{tabular}{l c}
\toprule
\textbf{Dataset Composition} & \textbf{Count} \\
\midrule
MedQA (train/dev) & 10178/1272 \\
MedMCQA (train) & 182822 \\
PubMedQA (train/val) & 450/50 \\\\
MMLU \\
Anatomy (dev/val) & 5/14 \\
Clinical Knowledge (dev/val) & 5/29 \\
College Biology (dev/val) & 5/16 \\
College Medicine (dev/val) & 5/22 \\
Medical Genetics (dev/val) & 5/11 \\
Professional Medicine (dev/val) & 5/31 \\
\midrule
\textbf{Total Count} & \textbf{194925} \\
\bottomrule
\end{tabular}
\end{table}

\subsection{Multi-Agent System for Complex CoT Generation}

\begin{figure}[ht]
    \centering
  \includegraphics[width=0.5\textwidth]{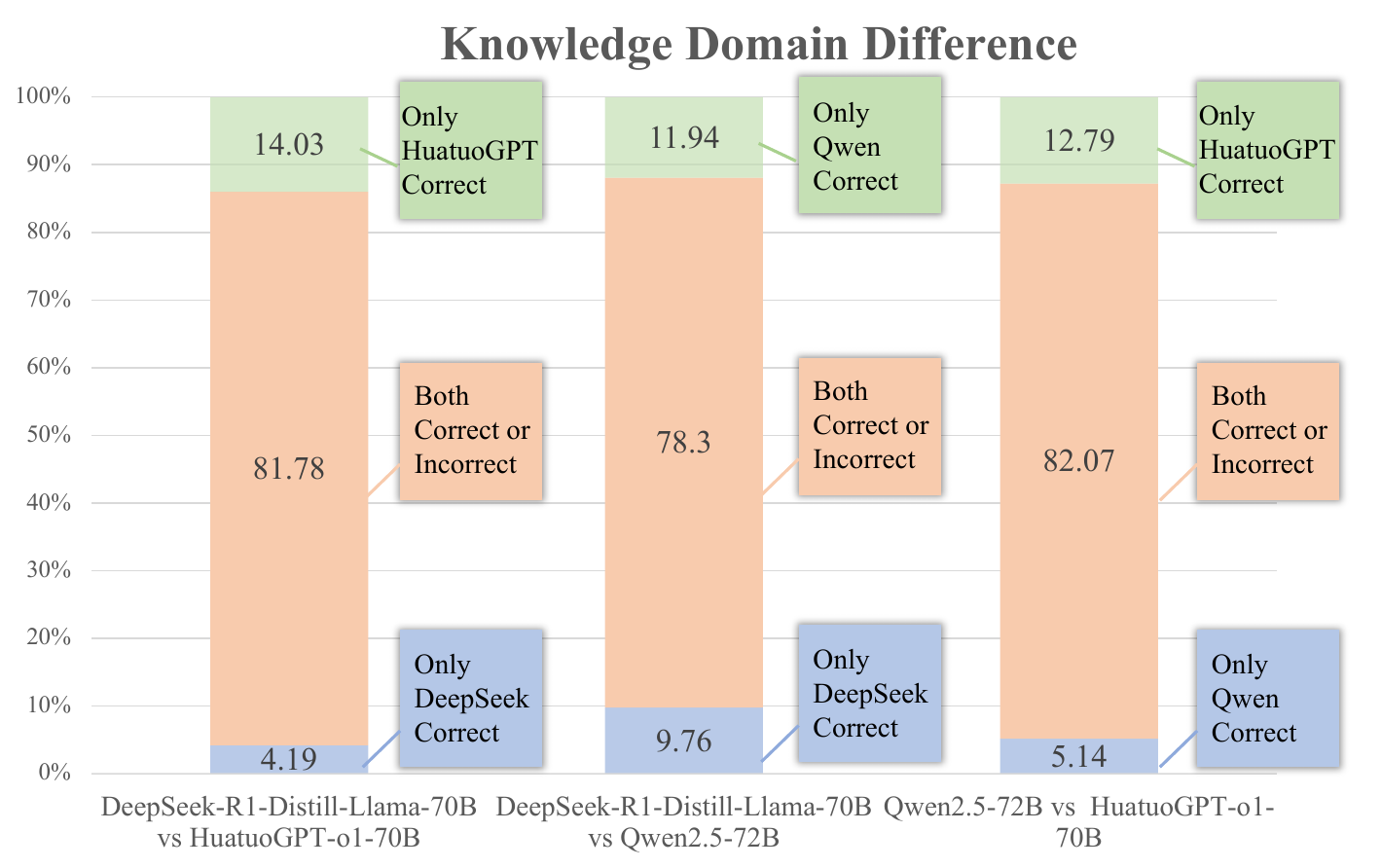}
  \caption{Knowledge domain differences among DeepSeek-R1-Distill-Llama-70B, HuatuoGPT-o1-70B and Qwen2.5-72B.}
  \label{fig:knowledge_diff}
\end{figure}

We employ a multi-agent framework\textemdash comprising Qwen2.5-72B, HuatuoGPT-o1-70B, and DeepSeek-R1-Distill-Llama-70B\textemdash to generate 1.755 million reasoning paths. Each model produces three CoT trajectories at different temperatures (0.7, 0.9, and 1.0). We then assemble the complex CoTs by following these steps:
\begin{enumerate}[label=(\roman*)]
    \item Rewrite the question.
    \item Highlight key clinical details and background information.
    \item Evaluate each answer choice and discuss supporting evidence and potential traps.
    \item Systematically eliminate choices inconsistent with the clinical context.
    \item Reassess each option, eliminating inconsistencies.
    \item Conclude with a final answer, supported by a concise explanation of the reasoning.
\end{enumerate}
As shown in Figure~\ref{fig:knowledge_diff}, we present a pairwise comparison among DeepSeek-R1-Distill-Llama-70B, HuatuoGPT-o1-70B, and Qwen2.5-72B on the Medical QA task. Specifically, we compare the number of questions correctly answered by each model individually. The results reveal that different models exhibit distinct strengths across various medical knowledge domains. The observed differences in knowledge domains across models highlight the necessity of a multi-agent system that integrates diverse model outputs.
In Fig. \ref {fig:knowledge_diff}, we present a pairwise comparison among DeepSeek-R1-Distill-Llama-70B, HuatuoGPT-o1-70B, and Qwen2.5-72B on the Medical QA task. Specifically, we compare the number of questions correctly answered by each model individually. The results reveal that different models exhibit distinct strengths across various medical knowledge domains.The observed differences in knowledge domains across models highlight the necessity of a multi-agent system that integrates diverse model outputs.

\subsection{Component Design}
\begin{figure*}[ht]
  \includegraphics[width=\textwidth]{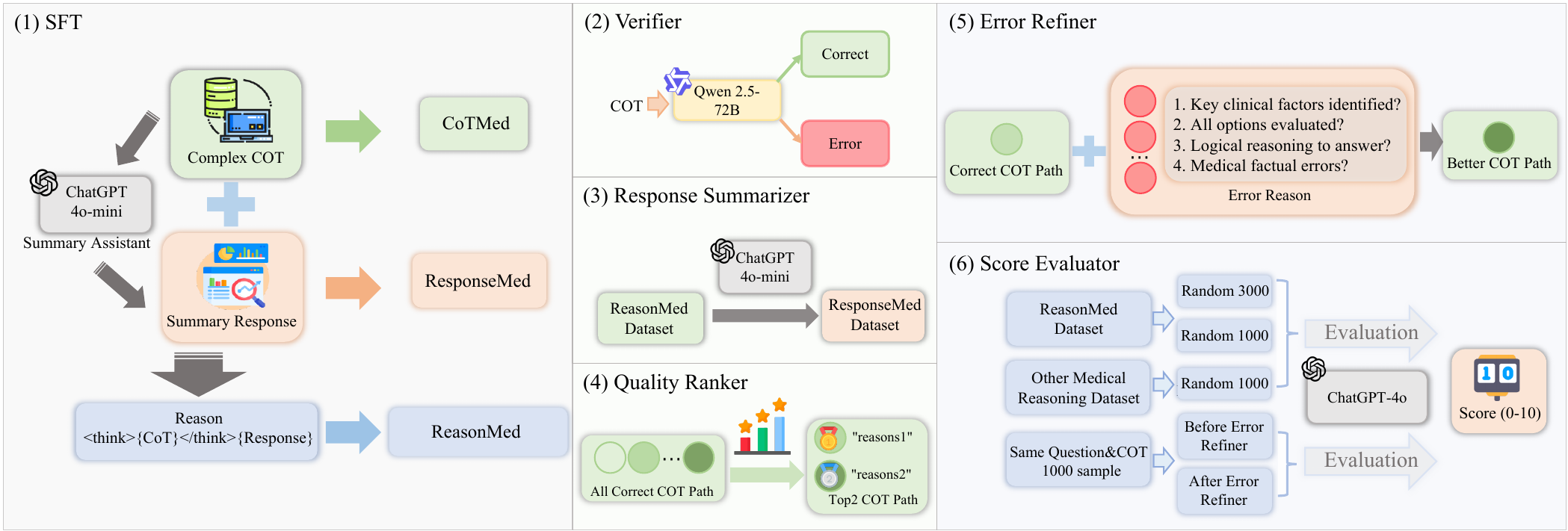}
  \caption{(1) Shows an example of SFT applied at different scales. (2) to (6) represent the components used to build the entire pipeline for our dataset.}
  \label{fig:Pipeline2}
\end{figure*}

This section provides an overview of the components developed in this paper and their respective functions. (2)-(6) of Figure~\ref {fig:Pipeline2} visualize the structure and workflow of each component.

\begin{figure}[ht]
\centering
  \includegraphics[width=0.7\columnwidth]{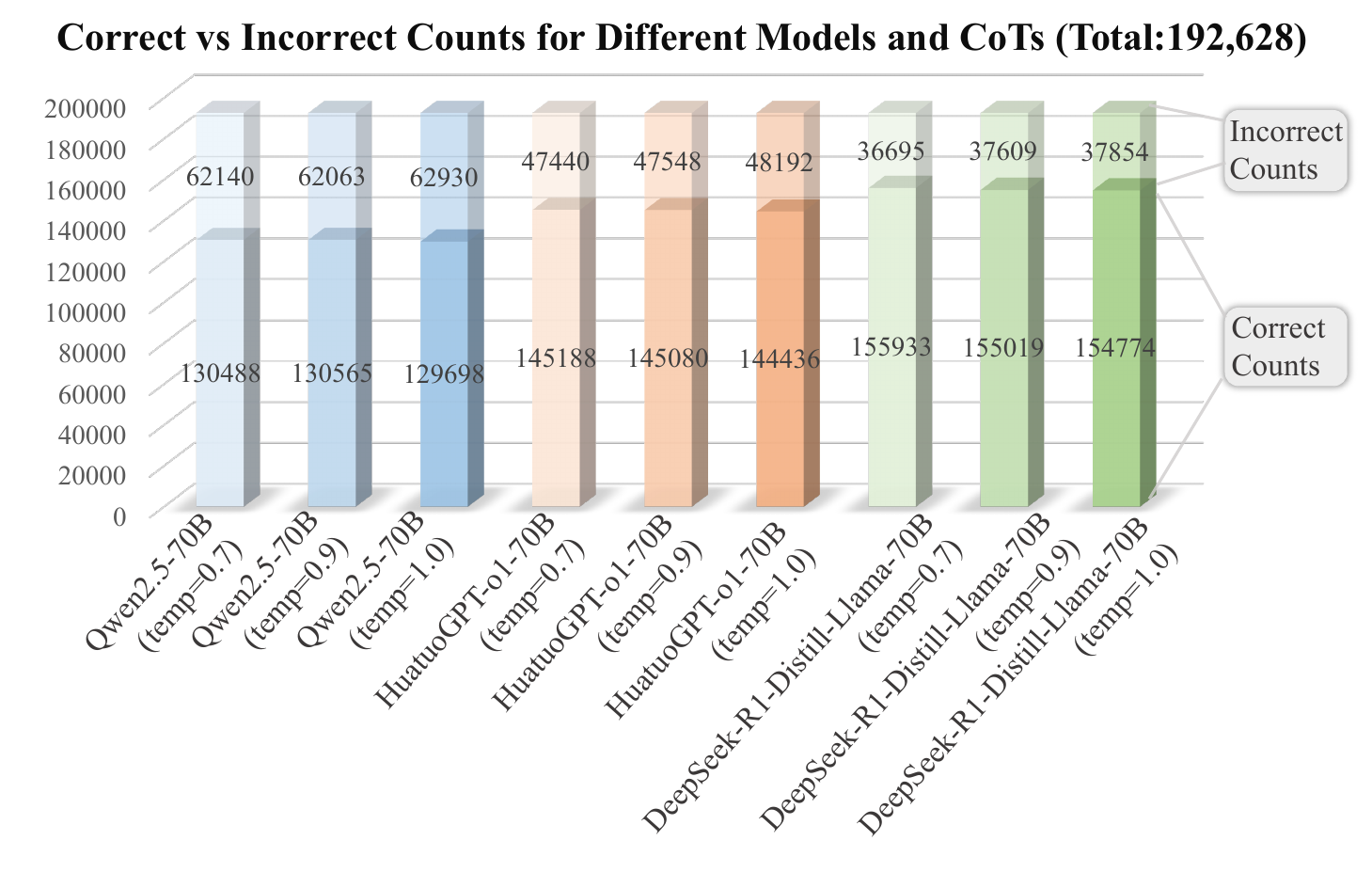}
  \caption{Bar chart illustrating the correct and incorrect counts for each model and CoT configuration across 9 generated paths in a Multi-Agent System, totaling 192,628.}
  \label{fig:CorrectvsIncorrect}
\end{figure}

\paragraph{Verifier:} This component constructs a verifier (based on Qwen2.5-72B) to validate the correctness of CoT paths generated by the Multi-Agent system. The model not only checks whether the answer is correct or incorrect, but also evaluates whether the key clinical factors have been accurately identified, whether all answer choices have been analyzed, and whether there are any factual errors in the medical knowledge. The model outputs a \texttt{JSON} object with two keys: one indicating the verdict (Correct or Error), and the other providing the reason for the error. For example, ``The CoT analysis contains inaccuracies regarding vasopressin's role in glycogenolysis and incorrectly dismisses oxytocin without full consideration of its potential regulatory effects.'' Figure~\ref {fig:CorrectvsIncorrect} presents a bar chart showing the number of correct versus incorrect reasoning paths\textemdash after Verifier validation\textemdash for each model and CoT configuration across the nine generated paths. DeepSeek-R1-Distill-Llama-70B achieves the highest overall accuracy; Qwen2.5-72B retains the most correct paths at a temperature of 0.9, while the optimal temperature for the other two models is $0.7$.

\paragraph{Response Summarizer:} To construct a response with reasoning similar to o1 answers, we use GPT-4o-mini as a summarization assistant. The model generates a summary for each complex CoT, which represents a step-by-step reasoning process. This summary is presented as the final output to the user, focusing on the reasoning aspect of the response.

\begin{figure}[ht]
    \centering
  \includegraphics[width=0.7\columnwidth, trim=0 0 20 0, clip]{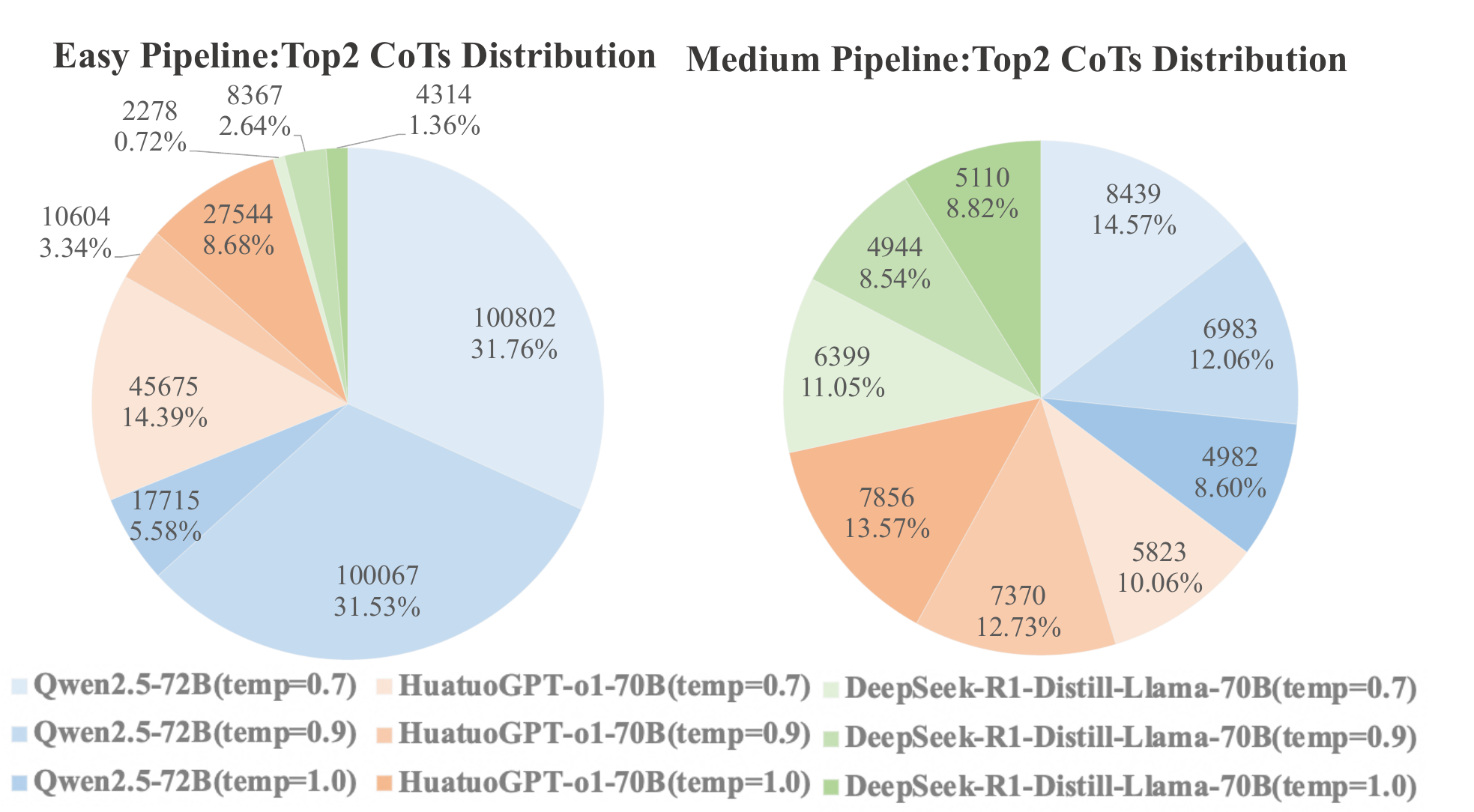}
  \caption{Distribution of the top two CoT paths selected by the Quality Ranker in Easy Pipeline and Medium Pipeline, showing sampling proportions across models and temperature settings.}
  \label{fig:Ranker}
\end{figure}

\paragraph{Quality Ranker:} Balancing dataset size and quality is crucial. Among the many correct CoT paths, we aim to select the two most optimal ones for subsequent training. The Quality Ranker, based on Qwen2.5-72B, plays a critical role here. The model reads the correct CoT paths and outputs the top two, such as ``top2'': [``modelX\_COTY'', ``modelZ\_COTW''], along with the rationale for excluding the other options. Initially, we considered using a Score Evaluator to rate each CoT, but this approach was challenging due to cases where multiple CoTs might have identical scores, making it difficult to select the best. Therefore, we opted for directly outputting the two best paths by their CoT names. 
Figure~\ref {fig:Ranker} shows the distribution of the top two CoT paths selected by the Quality Ranker in both Easy Pipeline and Medium Pipeline, illustrating the sampling proportions across different models and temperature settings.

\paragraph{Error Refiner:} This component handles questions of moderate difficulty. Using the Quality Ranker, it first selects the two most optimal reasoning paths (if only two chains of thought are correct, they are chosen by default), and then performs a secondary optimization. Concretely, the refiner applies \emph{targeted edits} to the flagged steps while \emph{preserving correct sub‐steps and conclusions}, thereby reducing unnecessary changes and mitigating drift from the original rationale. Its design also includes storing the model’s error reasons during the verification stage and leveraging a stronger model to supplement and address those weak points-an approach that effectively corrects the model’s error-prone knowledge. During this pass, the refiner prioritizes issues surfaced by the Verifier (\emph{e.g.}, missing clinical factors, incomplete option analysis, or factual slips) and amends only the implicated segments, followed by a brief self-consistency check to ensure the final answer and supporting evidence remain aligned. Empirically, Medium-pipeline refinement raises the Score-Evaluator average from 7.37 to 8.17 (+0.80), validating the Error Refiner's contribution. Overall, this localized, feedback-driven correction improves coherence and factual fidelity without overhauling well-formed reasoning, yielding higher-quality supervision signals for subsequent fine-tuning.

\paragraph{Score Evaluator:} This component utilizes the GPT-4o API to score the dataset quality on a scale from 0 to 10. We conducted two main experiments: the first compared the scores of the same question before and after CoT optimization to validate the effectiveness of the Error Refiner; the second involved comparing our final ReasonMed with other open-source medical reasoning datasets through random sampling to assess the effectiveness of our Multi-Agent approach.

\subsection{ReasonMed Build Pipeline} 

Based on the number of errors detected in the reasoning paths, three distinct pipelines were created to process CoTs at varying levels of difficulty:
\paragraph{Easy Pipeline (Error 0-4):} This pipeline handles paths with few errors (0-4), which are relatively easy for the model to answer correctly. Here, we use Quality Ranker to rank the correct paths, selecting the top two from the 5-9 correct options. Additionally, the model provides brief explanations as to why it did not choose other CoT paths.

\paragraph{Medium Pipeline (Error 5-7):} For paths with moderate errors (5-7), we assume that the model has partial knowledge but may miss certain fine-grained details. Thus, the top two CoT paths are selected using the Quality Ranker and then refined using the Error Refiner based on the pitfalls provided by the Verifier, focusing on correcting those errors to enhance the original correct reasoning chains.

\paragraph{Difficult Pipeline (Error 8-9):} For difficult questions with significant errors (8-9), the GPT-4o model may not be sufficient to correct the mistakes. Therefore, we use GPT-o1 to optimize these paths. For entirely incorrect paths, GPT-o1 generates high-quality CoTs from scratch, following the six-step reasoning process.

\begin{figure}[ht]
\centering
  \includegraphics[width=0.5\columnwidth, trim=150 0 140 0, clip]{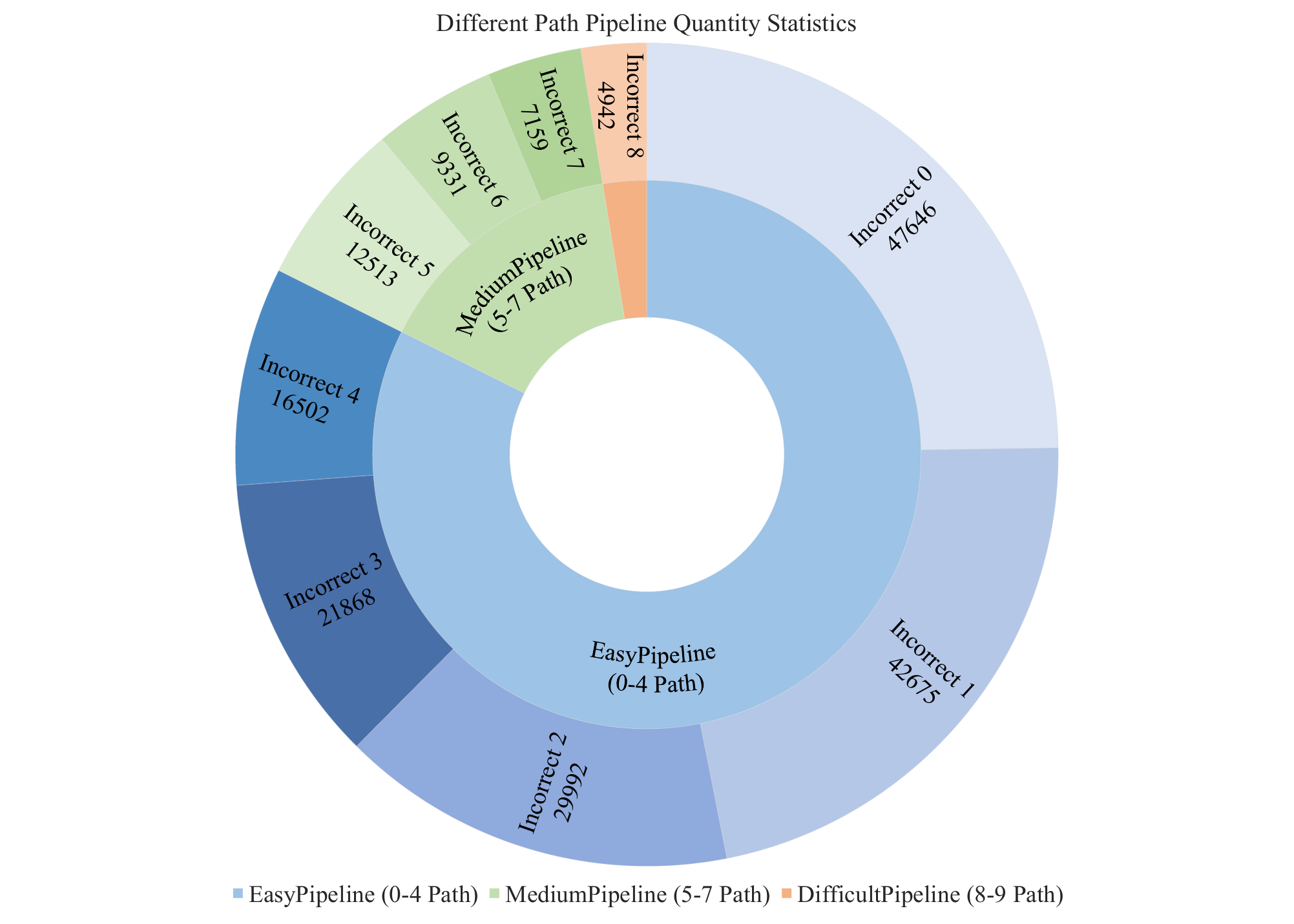}
  \caption{Different Pipeline Quantity Statistics.}
  \label{fig:pipeline_path_qualities}
\end{figure}

Lastly, Figure~\ref {fig:pipeline_path_qualities} presents the different pipeline quantity statistics, showing the distribution of paths handled by Easy, Medium, and Difficult pipelines.

By analyzing the number of correct paths validated by the Verifier, we can approximate each question’s difficulty. Accordingly, we design three distinct pipelines to tackle problems of varying complexity, systematically correcting errors in complex CoTs and refining the original dataset to strike an optimal balance between scale and quality.

\section{Multiscale Supervised Fine‐Tuning}
To assess the impact of explicit reasoning supervision on a downstream medical QA task, we propose a multiscale fine-tuning strategy leveraging three variants of our high-quality dataset. These variants are based on different granularities of reasoning, as outlined below:

\begin{itemize}
  \item \textbf{CoT:} A complex chain of thought consisting of six reasoning steps,
  \item \textbf{Response:} A concise response generated by a Response Summarizer from the CoT,
  \item \textbf{Reason:} A combination of the complex CoT and its corresponding summarized response.
\end{itemize}

\subsection{Data Preparation}
Leveraging the 370K ReasonMed introduced in Section~\ref{sec:multi_agent_reasoning_pipeline}, we employ a Response Summarizer to condense each chain-of-thought into a succinct answer explanation. For every question q and its corresponding CoT path \(Multi-step = [step_1, \dots, step_6]\), we generate the following instances:

\begin{itemize}
  \item \textbf{CoT instance:}  
    \[
      [\,q; step_1, step_2, \dots, step_6\,] \;\mapsto\; \texttt{CoT},
    \]
  \item \textbf{Response instance:}  
    \[
      \text{Response Summarizer}(\texttt{CoT}) \;\mapsto\; \texttt{Response},
    \]
  \item \textbf{Reason instance:}
    \[
      \texttt{<think>\{CoT\}</think>} \texttt{Response}\\
      \;\mapsto\; \texttt{Reason}.
    \]
\end{itemize}

The CoT, Response, and Reason instances are designed to encapsulate different levels of reasoning and summarization, providing a different scale of data for training.

\subsection{Fine-Tuning and Training}
We fine-tune the open-source Qwen2.5-7B model using three different fine-tuning regimes, with each regime corresponding to a different data scale. Specifically, we utilize LlamaFactory~\cite{zheng2024llamafactory} to perform 3 epochs of supervised fine-tuning on the following datasets:

\begin{itemize}
  \item \textbf{CoTMed-7B:} Fine-tuned with the CoT instances, focusing on reproducing the reasoning trace and generating the final answer.
  \item \textbf{ResponseMed-7B:} Fine-tuned with the Response instances, where the model is trained to generate concise summaries of the reasoning path.
  \item \textbf{ReasonMed-7B:} Fine-tuned with the Reason instances, combining detailed reasoning with summarized feedback.
\end{itemize}

Figure~\ref {fig:Pipeline2}(1) illustrates the SFT process. For evaluation, we use the \texttt{lm\_eval} framework to analyze the performance of these models on benchmark tasks, examining whether multi-step reasoning could enhance the model's ability to perform medical QA. We also train models with fewer epochs, including a variant trained for only one epoch, to assess performance differences and investigate the effect of fewer training steps. The results of these experiments will be discussed in detail in the experimental section.

\subsection{Training Details}
We perform full-model fine-tuning of the Qwen2.5-7B checkpoint using the LLaMA-Factory framework on a 16 x H20 GPU cluster. The ResponseMed configuration completed in approximately 9 hours, whereas CoTMed and ReasonMed required roughly 25 hours and 28 hours, respectively.

\section{Experiments}

\subsection{Dataset Quality Evaluation}

\paragraph{Medium Pipeline Validity Verification:}
To evaluate the effectiveness of the Medium Pipeline, we sample $1,000$ questions + CoT and use the Score Evaluator to assess the quality of answers both before and after applying the Medium Pipeline (GPT-4o-mini corrections). For each item, we conduct a paired evaluation using the same rubric and prompts, ensuring that pre- and post-optimization scores are directly comparable under identical conditions.

The results show a significant improvement, with an average score increase of $0.8$ points post-optimization. 
This paired setup isolates the contribution of the Error Refiner step, indicating that targeted edits improve coherence, factual fidelity, and option analysis without altering the underlying question distribution.
The specific scores are as follows:

\begin{table}[ht]
  \centering
  \caption{Score Evaluator results for Medium Pipeline validity.}
  \label{tab:data-quality}
    \begin{tabular}{lcc}
      \toprule
      \textbf{Dataset}                   & \textbf{Samples} & \textbf{Avg.\ Score} \\
      \midrule
      Medium Pipeline (pre-opt)          & 1,000            & 7.37                 \\
      Medium Pipeline (post-opt)         & 1,000            & 8.17                 \\
      \bottomrule
    \end{tabular}%
\end{table}

\paragraph{Comparison with Open-Source Datasets:}
We compare the ReasonMed with two publicly open-source medical reasoning corpora: \texttt{medical-o1-reasoning-SFT} and \texttt{Medical-R1-Distill-Data}. For a fair comparison, we sample $1,000$ instances from each of these datasets and extend ReasonMed with an additional $3,000$ samples. The results demonstrate that ReasonMed outperforms both baselines, achieving an average score of $8.45$ for the $1,000$ sample subset and $8.50$ for the $3,000$ sample subset. This represents an improvement of $3.9\%$ and $5.9\%$ over the other datasets, respectively.

\begin{table}[ht]
  \centering
  \caption{Score Evaluator results for comparison with other datasets.}
  \label{tab:data-quality2}
    \begin{tabular}{lcc}
      \toprule
      \textbf{Dataset}                   & \textbf{Samples} & \textbf{Avg.\ Score} \\
      \midrule
      medical-o1-reasoning-SFT           & 1,000            & 8.03                 \\
      Medical-R1-Distill-Data            & 1,000            & 8.18                 \\
      ReasonMed                          & 1,000            & 8.45                 \\
      ReasonMed                          & 3,000            & 8.50                 \\
      \bottomrule
    \end{tabular}%
\end{table}

\subsection{Multiscale Supervised Fine-Tuning}

In this section, we present a comprehensive analysis of the experimental results obtained by fine-tuning the Qwen2.5-7B model using our proposed multiscale supervised fine-tuning (SFT) strategy. Performance comparisons across various medical question-answering (QA) benchmarks, including MedQA, MedMCQA, PubMedQA, and MMLU, are detailed in Table~\ref {table:performance_comparison}.
Our results demonstrate the effectiveness of incorporating explicit reasoning supervision at multiple granularities:

\paragraph{CoTMed-7B} consistently outperforms baseline models across most benchmarks, achieving notably higher scores in MedQA ($66.3\%$), MedMCQA ($64.7\%$), and PubMedQA ($80.0\%$). This indicates that fine-tuning on complex reasoning chains substantially enhances the model's capacity to perform medical reasoning tasks.

\paragraph{ResponseMed-7B} focuses solely on generating concise summaries, achieving competitive results, with notable performance on MedQA ($67.5\%$) but slightly lower overall accuracy ($67.0\%$) compared to CoTMed-7B ($69.1\%$). This suggests that while response summarization captures key information effectively, it may miss nuanced reasoning steps critical for complex questions.

\paragraph{ReasonMed-7B} combines detailed reasoning chains and concise summaries, yielding the highest total accuracy ($69.6\%$), particularly excelling in MedMCQA ($65.1\%$) and PubMedQA ($82.0\%$). This hybrid approach appears to effectively leverage the strengths of both granularities, achieving balanced and robust performance across diverse question types.

To explore the impact of training duration, we also compare model performances trained for different epochs:
\paragraph{One Epoch Training:} Models trained for one epoch showed promising yet suboptimal performance compared to their three-epoch counterparts. CoTMed-1epoch achieved an overall accuracy of $67.8\%$, slightly outperforming ReasonMed-7B-1epoch ($67.7\%$) and significantly surpassing ResponseMed-7B-1epoch ($64.8\%$).
\paragraph{Three Epoch Training:} Models trained for three epochs consistently improved across benchmarks, clearly illustrating the benefit of extended training, with overall accuracy improved from $67.71\%$ (1 epoch) to $69.63\%$ (3 epochs).

Under limited training steps, CoTMed-7B outperforms ReasonMed-7B. However, as the number of training steps increases, ReasonMed-7B ultimately surpasses CoTMed-7B by $0.54\%$. Additional training may enable the model to more effectively learn the internal connections between complex CoT reasoning and concise summarization, resulting in further performance gains.

\begin{table*}[ht!] \small \centering
\adjustbox{max width=1.0\textwidth}{
\begin{tabular}{lccccccccccc} \toprule 
& \multirow{3}{*}{\textbf{MedQA}} & \multirow{3}{*}{\textbf{MedMCQA (val)}} & \multirow{3}{*}{\textbf{PubMedQA}}  & \multicolumn{6}{c}{\textbf{MMLU}} & \multirow{3}{*}{\textbf{Total Acc}} & \multirow{3}{*}{\textbf{Avg. token}} \\ 
\cmidrule(r){5-10} 
& & & & \multirow{2}{*}{\textbf{Anatomy}}  &   \multirow{2}{*}{\makecell{\textbf{CK}}} & \multirow{2}{*}{\makecell{\textbf{C-Bio}}} & \multirow{2}{*}{\makecell{\textbf{C-Med}}} & \multirow{2}{*}{\makecell{\textbf{Med-Gene}}} & \multirow{2}{*}{\makecell{\textbf{P-Med}}} && \\ 
&&&&&&&&&&\\
\midrule
\textbf{Dataset\_Count} & 1273 & 4183 & 1000 & 135 & 265 & 144 & 173 & 100 & 272 & - & - \\
\midrule
\multicolumn{12}{l}{\textbf{ Models $<$ 10B}} \\ \midrule
\addlinespace[2pt]
BioMistral-7B & $45.6 \pm 1.4$ & $41.5 \pm 0.8$ & $71.0 \pm 2.0$ & $\underline{76.3 \pm 3.7}$ & $63.0 \pm 3.0$ & $62.5 \pm 4.1$ & $53.8 \pm 3.8$ & $67.0 \pm 4.7$ & $53.3 \pm 3.0$ & 48.9 & 60.1 \\
Llama3-OpenBioLLM-8B & $57.9 \pm 1.4$ & $57.7 \pm 0.8$ & $76.0 \pm 6.1$ & $68.9 \pm 4.0$ & $77.7 \pm 2.6$ & $83.3 \pm 3.1$ & $69.4 \pm 3.5$ & $83.0 \pm 3.8$ & $79.0 \pm 2.5$ & 62.9 & 75.1 \\
Llama-3-8B-UltraMedical & $63.2 \pm 1.4$ & $57.7 \pm 0.8$ & $78.0 \pm 5.9$ & $67.4 \pm 4.1$ & $74.3 \pm 2.7$ & $75.7 \pm 3.6$ & $61.9 \pm 3.7$ & $73.0 \pm 4.5$ & $78.7 \pm 2.5$ & 63.5 & 5177.7 \\
Mistral-7B-Instruct-v0.3 & $52.2 \pm 1.4$ & $48.2 \pm 0.8$ & $82.0 \pm 5.5$ & $59.3 \pm 4.2$ & $69.4 \pm 2.8$ & $72.9 \pm 3.7$ & $56.7 \pm 3.8$ & $70.0 \pm 4.6$ & $66.5 \pm 2.9$ & 55.9 & 111.8 \\
Yi-1.5-9B-Chatbot & $49.8 \pm 1.4$ & $47.0 \pm 0.8$ & $69.0 \pm 2.1$ & $67.5 \pm 3.8$ & $63.9 \pm 2.8$ & $70.3 \pm 3.8$ & $51.2 \pm 4.0$ & $68.8 \pm 4.5$ & $66.7 \pm 3.1$ & 52.9 & 162.2 \\
HuatuoGPT-o1-7B & $\mathbf{68.4 \pm 1.3}$ & $57.5 \pm 0.8$ & $74.0 \pm 2.0$ & $71.9 \pm 3.9$ & $78.5 \pm 2.5$ & $\mathbf{88.2 \pm 2.7}$ & $67.6 \pm 3.6$ & $80.0 \pm 4.0$ & $77.6 \pm 2.5$ & 64.4 & 446.0 \\
HuatuoGPT-o1-8B & $65.4 \pm 1.3$ & $61.0 \pm 0.8$ & $74.6 \pm 2.0$ & $69.6 \pm 4.0$ & $77.7 \pm 2.6$ & $81.3 \pm 3.3$ & $69.9 \pm 3.5$ & $78.0 \pm 4.2$ & $71.0 \pm 2.8$ & 65.5 & 468.9 \\
\addlinespace[2pt]
ResponseMed-7B (1epoch) & $62.2 \pm 1.4$ & $57.6 \pm 0.8$ & $\mathbf{84.0 \pm 5.2}$ & $75.6 \pm 3.7$ & $77.7 \pm 2.6$ & $81.3 \pm 3.3$ & $69.9 \pm 3.5$ & $87.0 \pm 3.4$ & $76.8 \pm 2.6$ & 64.8 & - \\
CoTMed-7B (1epoch) & $64.3 \pm 1.3$ & $62.4 \pm 0.8$ & $\underline{82.0 \pm 5.5}$ & $\mathbf{77.0 \pm 3.6}$ & $\mathbf{80.8 \pm 2.4}$ & $81.3 \pm 3.3$ & $72.8 \pm 3.4$ & $\mathbf{90.0 \pm 3.0}$ & $79.4 \pm 2.5$ & 67.8 & - \\
ReasonMed-7B (1epoch) & $65.3 \pm 1.3$ & $62.3 \pm 0.8$ & $\underline{82.0 \pm 5.5}$ & $74.8 \pm 3.7$ & $\underline{80.0 \pm 2.5}$ & $81.3 \pm 3.3$ & $\mathbf{74.0 \pm 3.4}$ & $86.0 \pm 3.5$ & $79.0 \pm 2.5$ & 67.7 & - \\
\addlinespace[2pt]
ResponseMed-7B & $\underline{67.5 \pm 1.3}$ & $60.9 \pm 0.8$ & $80.0 \pm 5.7$ & $74.8 \pm 3.7$ & $77.4 \pm 2.6$ & $\underline{84.0 \pm 3.1}$ & $71.1 \pm 3.5$ & $\underline{88.0 \pm 3.3}$ & $76.5 \pm 2.6$ & 67.0 & 225.2 \\
CoTMed-7B & $66.3 \pm 1.3$ & $\underline{64.7 \pm 0.7}$ & $80.0 \pm 5.7$ & $75.6 \pm 3.7$ & $79.6 \pm 2.5$ & $82.1 \pm 3.2$ & $71.7 \pm 3.4$ & $86.0 \pm 3.5$ & $\underline{79.9 \pm 2.6}$ & $\underline{69.1}$ & 555.4 \\
ReasonMed-7B & $66.9 \pm 1.3$ & $\mathbf{65.1 \pm 0.7}$ & $\underline{82.0 \pm 5.5}$ & $75.6 \pm 3.7$ & $79.3 \pm 2.5$ & $79.2 \pm 3.4$ & $\underline{73.4 \pm 3.4}$ & $85.0 \pm 3.6$ & $\mathbf{80.9 \pm 2.4}$ & $\mathbf{69.6}$ & 626.0 \\
\midrule
\multicolumn{12}{l}{\textbf{Scaling: Models $>$ 10B}} \\ \midrule
\addlinespace[2pt]
LLaMA3.1-70B & $76.8 \pm 0.1$ & $67.9 \pm 0.7$ & $77.4 \pm 0.2$ & $81.5 \pm 0.3$ & $89.1 \pm 0.2$ & $96.5 \pm 0.1$ & $80.9 \pm 0.3$ & $90.0 \pm 0.3$ & $93.0 \pm 0.2$ & $\mathbf{72.9}$ & - \\
Qwen2.5-14B & $75.6 \pm 0.1$ & $63.4 \pm 0.8$ & $75.6 \pm 0.4$ & $75.6 \pm 0.4$ & $84.9 \pm 0.2$ & $88.9 \pm 0.3$ & $75.7 \pm 0.3$ & $90.0 \pm 0.3$ & $84.2 \pm 0.2$ & 69.0 & - \\
Qwen2.5-32B & $79.3 \pm 0.1$ & $67.6 \pm 0.7$ & $77.6 \pm 0.2$ & $79.3 \pm 0.3$ & $86.8 \pm 0.2$ & $93.8 \pm 0.2$ & $79.8 \pm 0.3$ & $91.0 \pm 0.3$ & $87.5 \pm 0.2$ & 72.6 & - \\
QwQ-32B & $78.1 \pm 0.1$ & $65.5 \pm 0.7$ & $76.4 \pm 0.2$ & $75.6 \pm 0.4$ & $86.8 \pm 0.2$ & $93.8 \pm 0.2$ & $77.5 \pm 0.3$ & $92.0 \pm 0.3$ & $88.2 \pm 0.2$ & 72.0 & - \\
CoTMed-14B & $73.5 \pm 1.2$ & $66.7 \pm 0.7$ & $80.0 \pm 5.7$ & $72.6 \pm 3.9$ & $83.0 \pm 2.3$ & $88.2 \pm 2.7$ & $78.6 \pm 3.1$ & $86.0 \pm 3.5$ & $84.9 \pm 2.2$ & 71.9 & - \\
ReasonMed-14B & $74.2 \pm 1.2$ & $67.6 \pm 0.7$ & $82.0 \pm 5.5$ & $74.1 \pm 3.8$ & $83.0 \pm 2.3$ & $88.9 \pm 2.6$ & $76.3 \pm 3.2$ & $86.0 \pm 3.5$ & $86.4 \pm 2.1$ & $\mathbf{72.8}$ & - \\
\bottomrule
\end{tabular}
}
\caption{Performance comparison of various models on MedQA, MedMCQA, PubMedQA, and MMLU benchmarks with total accuracy and average token length, where CK, C-Bio, C-Med, Med-Gene, and P-Med denote Clinical Knowledge, College Biology, College Medicine, Medical Genetics, and Professional Medicine, respectively. We present two clearly separated blocks: \emph{Models $<$ 10B} and \emph{Scaling: Models $>$ 10B}.}
\label{table:performance_comparison}
\end{table*}

\paragraph{Analysis of Average Token Length:}  
To compute these averages, each model was evaluated in inference mode across the entire test set, and the mean number of generated tokens was recorded. CoTMed-7B ($\approx555$ tokens) and ReasonMed-7B ($\approx626$ tokens) consistently produced substantially longer outputs than ResponseMed-7B ($\approx225$ tokens), suggesting a more elaborate reasoning process at the expense of conciseness. Relative to HuatuoGPT-o1-7B ($\approx446$ tokens), our CoTMed and ReasonMed variants demonstrate even more expansive chains of thought. Notably, although ResponseMed-7B generates significantly fewer tokens, it still surpasses the HuatuoGPT-o1 models in overall accuracy, underscoring the decisive role of dataset scale and quality in shaping model performance.

Compared with other biomedical LLMs such as BioMistral-7B, Llama3-OpenBioLLM-8B, and HuatuoGPT-o1, our ReasonMed-7B exhibits superior performance on medical QA tasks, attaining the highest aggregate metrics across benchmarks. It surpasses the strongest model of comparable scale by $4.17\%$, and even exceeds the performance of certain ten-billion-parameter systems on multiple evaluations (see Appendix). These findings highlight the critical role of both dataset quality and scale, while further emphasizing the effectiveness of explicit multi-step reasoning in advancing medical QA. Moreover, with extended training iterations, ReasonMed-7B increasingly internalizes the linkage between elaborate reasoning chains and succinct answer formulations, thereby yielding notable gains in overall capability.

\paragraph{Scaling Beyond 7B:}
To rigorously evaluate scalability under practical computational constraints, we expand our investigation from 7B to 14B parameter backbones while preserving an identical training protocol. On nine medical benchmarks—including MedQA, MedMCQA,(val), PubMedQA, and six MMLU medical sub-domains (Anatomy, Clinical Knowledge, College Biology, College Medicine, Medical Genetics, and Professional Medicine)—ReasonMed-14B attains an overall accuracy of $72.8\%$. This performance reflects a $+3.8\%$ gain over Qwen2.5-14B ($69.0\%$), surpasses Qwen2.5-32B ($72.6\%$), and nearly matches LLaMA3.1-70B ($72.9\%$). These findings underscore that our data-centric paradigm delivers consistent scaling improvements and exceptional data efficiency at the 14B scale, substantially narrowing the gap to considerably larger models while avoiding their prohibitive training costs.

\paragraph{Implementation, Cost, and Decoding Summary (see Appendix): }
For reproducibility and practical budgeting, we provide in the Appendix a consolidated set of implementation details. Specifically, we report: (i) a compute profile covering data generation, training, and inference (\ref{app:costs}, Table~\ref{tab:train_infer_profile}); (ii) the complete training and default inference hyperparameters (\ref{app:hyperparams}, Table~\ref{tab:train_hparams}); (iii) a decoding sweep demonstrating that accuracy stabilizes at temperatures within $[0.5,0.7]$ (\ref{app:temp_sweep}, Table~\ref{tab:temp_sweep}); and (iv) a cost analysis comparing direct GPT-o1 distillation against our selective EMD pipeline, which achieves a $\sim$3.6$\times$ reduction in expense (\ref{app:emd_cost}, Table~\ref{tab:emh_vs_o1}). Full numerical results and operational specifications are included in the cited Appendix sections.

\section{Conclusion}
In this work, we present \textbf{ReasonMed}, the largest open-source dataset dedicated to medical reasoning, developed to advance model performance on complex medical question answering tasks. Leveraging a multi-agent framework, we systematically generate, validate, and refine 1.291 million reasoning trajectories, distilling them into 370k high-quality exemplars. Extensive experiments confirm that explicit multi-step reasoning yields substantial performance gains, with our hybrid paradigm, integrating Chain-of-Thought reasoning and summarization, achieving state-of-the-art results. Remarkably, ReasonMed-7B/14B models consistently surpass strong baselines, even outperforming counterparts with significantly larger parameter scales. These results highlight the pivotal role of reasoning in medical QA and establish a scalable, data-centric methodology for future research in knowledge-intensive domains. Beyond the medical field, our framework offers a transferable blueprint for constructing domain-specific reasoning datasets, laying the foundation for broader applications across diverse scientific and technical disciplines.

\section*{Limitations}
While we extend our study beyond 7B to a 14B backbone and report competitive results against 14B/32B and even 70B models on multiple benchmarks, our analysis still has several limitations. 
We have not conducted full ablations for models above 14B (\emph{e.g.}, 32B/70B) under identical training recipes. Thus, the scalability of our data-centric approach at frontier scales remains partially assessed. 
Portions of our filtering (Verifier, Quality Ranker) and automatic quality assessment (Score Evaluator) rely on strong LLM judges (Qwen2.5-72B, GPT-4o). Despite pilot physician review, such reliance can propagate judge biases and systematic errors. 

%

\bibliography{main}
\bibliographystyle{icml2025}


\newpage
\appendix
\renewcommand{\thesection}{\Alph{section}}

\startcontents[appendix]
\section*{Appendix}
\section*{Appendix Contents}
\printcontents[appendix]{}{1}{%
  \setcounter{tocdepth}{2}
  \renewcommand{\thesection}{\Alph{section}}
  \renewcommand{\thesubsection}{\Alph{section}.\arabic{subsection}}
}

\section{Ethical Statement}
\label{app:ethics}
The ReasonMed-7B/14B model presented in this paper has demonstrated strong performance in handling complex medical reasoning tasks. Nonetheless, it still carries a risk of generating inaccurate information, incomplete explanations, or hallucinations, which could potentially mislead users. Therefore, we strongly advise against the direct use of this model in clinical settings or any real-world applications where errors might lead to significant negative consequences. To ensure responsible usage, we restrict the model exclusively to academic research purposes. It is essential for users to recognize and respect these guidelines, thus avoiding situations in which the dissemination of incorrect medical information could compromise patient safety, treatment accuracy, or clinical judgment.

\definecolor{TiffanyBlue}{RGB}{129,216,207}
\section{Component Prompt Design}
\subsection{CoT Generate}
This component is used to generate medical MCQ analysis prompts with detailed chain thinking (CoT) to guide the model for step-by-step reasoning.

\begin{tcolorbox}[breakable, colback=TiffanyBlue!5!white, colframe=TiffanyBlue!75!black, title={CoT Generate}]
\begin{lstlisting}[
    numbers=none, 
    frame=none,
    basicstyle=\ttfamily\small\fontseries{l}\selectfont,
    columns=fullflexible,
    breaklines=true,
    breakatwhitespace=true,             
    prebreak={\mbox{\thinspace}},       
    postbreak={},                       
    xleftmargin=0pt,                    
    xrightmargin=0pt,                   
    breakindent=0pt                  
]
"""
You are a highly knowledgeable medical expert. You are provided with a clinical multiple-choice question along with several candidate answers.
Your task is to carefully analyze the clinical scenario and each option by following these steps:
1. Restate the question in your own words.
2. Highlight the key clinical details and relevant background information (e.g., pathophysiology, anatomy, typical presentations, diagnostic tests).
3. Evaluate each candidate answer, discussing supporting evidence and potential pitfalls.
4. Systematically rule out options that do not align with the clinical context.
5. Compare any remaining choices based on their merits.
6. Conclude with your final answer accompanied by a clear and concise summary of your reasoning.

Please note: Your response should be based solely on the current question and candidate answers. Do not consider any previous context or prior interactions.

Question:
{question}

Candidate Answers:
{options}

Please provide your detailed chain-of-thought reasoning followed by your final answer.
"""
\end{lstlisting}
\end{tcolorbox}

\subsection{Verifier}
This component is used to evaluate the chain-of-thoughts generated by the Multi-Agent system to determine whether their reasoning is correct and output JSON results.
\begin{tcolorbox}[breakable, colback=TiffanyBlue!5!white, colframe=TiffanyBlue!75!black, title={Verifier}]
\begin{lstlisting}[
    numbers=none, 
    frame=none,
    basicstyle=\ttfamily\small\fontseries{l}\selectfont,
    columns=fullflexible,
    breaklines=true,
    breakatwhitespace=true,             
    prebreak={\mbox{\thinspace}},       
    postbreak={},                       
    xleftmargin=0pt,                    
    xrightmargin=0pt,                   
    breakindent=0pt                  
]
"""
You are a medical evaluation expert. Analyze if the Chain-of-Thought (CoT) analysis correctly leads to the answer.

[Question]
{question}

[Options]
{options_str}

[Correct Answer]
{answer}

[CoT Analysis]
{cot_content}

Evaluate the CoT analysis following these criteria:
1. Does the analysis correctly identify key clinical factors?
2. Are all options appropriately considered and evaluated?
3. Does the reasoning logically lead to the correct answer?
4. Are there any factual errors in medical knowledge?

Output a JSON object with:\\
- "verdict": "Correct" if the CoT analysis is valid and reaches the correct answer, otherwise "Error"
- "reason": Brief explanation of your evaluation (1-2 sentences)
"""
\end{lstlisting}
\end{tcolorbox}

\subsection{Response Summarizer}
This component is used to refine long-form CoT reasoning into concise summaries.
\begin{tcolorbox}[breakable, colback=TiffanyBlue!5!white, colframe=TiffanyBlue!75!black, title={Response Summarizer}]
\begin{lstlisting}[
    numbers=none, 
    frame=none,
    basicstyle=\ttfamily\small\fontseries{l}\selectfont,
    columns=fullflexible,
    breaklines=true,
    breakatwhitespace=true,             
    prebreak={\mbox{\thinspace}},       
    postbreak={},                       
    xleftmargin=0pt,                    
    xrightmargin=0pt,                   
    breakindent=0pt                  
]
"""
Summarize the following chain-of-thought reasoning:
{cot}
"""
\end{lstlisting}
\end{tcolorbox}

\subsection{Quality Ranker}
This component is used to sort the correct chain of thought after Verifier is filtered by quality.
\begin{tcolorbox}[breakable, colback=TiffanyBlue!5!white, colframe=TiffanyBlue!75!black, title={Quality Ranker}]
\begin{lstlisting}[
    numbers=none, 
    frame=none,
    basicstyle=\ttfamily\small\fontseries{l}\selectfont,
    columns=fullflexible,
    breaklines=true,
    breakatwhitespace=true,             
    prebreak={\mbox{\thinspace}},       
    postbreak={},                       
    xleftmargin=0pt,                    
    xrightmargin=0pt,                   
    breakindent=0pt                  
]
"""
You are a medical reasoning evaluator. Given the question, options, and known answer, review the following chains-of-thought (CoTs) labeled by their keys.
Select the two most sound and useful CoTs, then provide brief justifications for why each of the other CoTs were not chosen.

[Question]
{question}

[Options]
A) {optA}
B) {optB}
C) {optC}
D) {optD}

[Correct Answer]
{answer}

[CoTs]
{cot_block}

Respond with a JSON object with exactly two keys:
  "top2": ["modelX_COTY", "modelZ_COTW"],
  "reasons": {<label>: <one-sentence justification> for every CoT not in top2}
"""
\end{lstlisting}
\end{tcolorbox}

\subsection{Error Refiner}
The Error Refiner improves moderate-difficulty items by selecting the top two CoT paths via the Quality Ranker and applying verifier-guided, targeted edits—leveraging a stronger model when needed—to correct factual/logical issues while preserving valid reasoning, thereby boosting coherence and producing higher-quality supervision for fine-tuning.
\begin{tcolorbox}[breakable, colback=TiffanyBlue!5!white, colframe=TiffanyBlue!75!black, title={Error Refiner}]
\begin{lstlisting}[
    numbers=none, 
    frame=none,
    basicstyle=\ttfamily\small\fontseries{l}\selectfont,
    columns=fullflexible,
    breaklines=true,
    breakatwhitespace=true,             
    prebreak={\mbox{\thinspace}},       
    postbreak={},                       
    xleftmargin=0pt,                    
    xrightmargin=0pt,                   
    breakindent=0pt                  
]
"""
You are an expert clinician-educator AI tutor. Your mission is to generate an exceptionally comprehensive, in-depth chain-of-thought explanation that rigorously justifies the correct answer for the given clinical MCQ, while specifically addressing and integrating provided error feedback to eliminate previous reasoning flaws. Adhere closely to these instructions to maximize completeness:

1. **Error-Driven Refinement**  
   - Review the provided **Error Reasons from Other Attempts**.  
   - Identify logical gaps, factual mistakes, omissions, or misleading inferences in the original chain‐of‐thought.  
   - Explicitly incorporate corrections and clarifications derived from these error reasons.

2. **Structured, Layered Reasoning**  
   Organize your explanation into clear sections:
   a. Restate the question in your own words.
   b. Highlight the key clinical details and relevant background information (e.g., pathophysiology, anatomy, typical presentations, diagnostic tests).
   c. Evaluate each candidate answer, discussing supporting evidence and potential pitfalls.
   d. Systematically rule out options that do not align with the clinical context.
   e. Compare any remaining choices based on their merits.
   f. Conclude with your final answer accompanied by a clear and concise summary of your reasoning.

**Inputs**
- **Question:**  '{question}'  
- **Options:**  '{options}'  
- **Correct Answer:**  '{answer}'  
- **Original Chain-of-Thought:**  '{original_cot}'  
- **Error Reasons from Other Attempts:**  '{error_reasons}'  

**Output:**  
Please optimized Original Chain-of-Thought. Ensure that you explicitly address and rectify each error reason provided.
"""
\end{lstlisting}
\end{tcolorbox}

\subsection{Score Evaluator}
This component is used to refine long-form CoT reasoning into concise summaries.
\begin{tcolorbox}[breakable, colback=TiffanyBlue!5!white, colframe=TiffanyBlue!75!black, title={Score Evaluator}]
\begin{lstlisting}[
    numbers=none, 
    frame=none,
    basicstyle=\ttfamily\small\fontseries{l}\selectfont,
    columns=fullflexible,
    breaklines=true,
    breakatwhitespace=true,             
    prebreak={\mbox{\thinspace}},       
    postbreak={},                       
    xleftmargin=0pt,                    
    xrightmargin=0pt,                   
    breakindent=0pt                  
]
"""
You are a medical reasoning evaluator. Assess the following response based on the following criteria:

1. **Clinical accuracy**: Does the response correctly incorporate medical facts, clinical guidelines, and evidence-based practices? Are the clinical details provided accurate, relevant, and appropriate for the given situation?
2. **Logical reasoning**: Does the response logically follow the reasoning process required to arrive at the answer? Is the reasoning chain coherent and well-supported by evidence or clinical knowledge?
3. **Factual correctness**: Are there any factual errors in the response? Are all statements factually correct and consistent with established medical knowledge?
4. **Completeness**: Does the response cover all necessary aspects of the question? Is it thorough and detailed, addressing the key points without missing critical information?

[Question]
{question}

[Response]
{response}

Please evaluate the response on the above criteria and provide a JSON object with two keys:
  "score": integer between 1 and 10,
  "justification": A concise explanation of your score.
"""
\end{lstlisting}
\end{tcolorbox}

\section{Training and Inference Compute Profile}
\label{app:costs}

Table~\ref{tab:train_infer_profile} summarizes the wall-clock compute for data generation, training, and inference of \textsc{ReasonMed}\textendash7B. These numbers complement the pipeline description in the main text and help estimate practical budgets.

\begin{table*}[ht]
  \centering
  \resizebox{\columnwidth}{!}{
  \begin{tabular}{l l l l}
  \toprule
  \textbf{Stage} & \textbf{What was done} & \textbf{Duration / Hardware} & \textbf{Notes} \\
  \midrule
  Data generation &
  \begin{tabular}[t]{@{}l@{}}
  1.75M reasoning paths with Qwen2.5\textendash72B,\\
  HuatuoGPT\textendash o1\textendash 70B, DeepSeek\textendash R1\textendash Distill\textendash Llama\textendash 70B\\
  Path validation with Qwen 2.5-72B \\ Ranking \& refinement with Easy + Medium pipelines
  \end{tabular}
  & $\approx 122$ h (end-to-end) & End-to-end pipeline runtime \\
  \addlinespace[2pt]
  Training & Full-parameter SFT for \textsc{ReasonMed}\textendash7B & $448$ GPU-hours (H20) & See Table~\ref{tab:train_hparams} \\
  \addlinespace[2pt]
  Inference & Serve \textsc{ReasonMed}\textendash7B & $1\times$ H20 GPU (fp16, bs=1) & Avg.\ response length $\approx 626$ tokens \\
  \bottomrule
  \end{tabular}}
  \caption{Compute profile for data generation, training, and inference serving.}
  \label{tab:train_infer_profile}
\end{table*}

\paragraph{Interpretation.}
Training dominates the \emph{model-side} compute. Data generation time reflects the aggregate of multi-agent sampling, verification, ranking, and medium-path refinement.

\section{Complete Training and Default Inference Hyperparameters}
\label{app:hyperparams}

Table~\ref{tab:train_hparams} lists the hyperparameters referenced in the supplementary notes (all other settings follow \texttt{LLaMA-Factory} defaults).

\begin{table*}[ht]
  \centering
  \begin{tabular}{l l}
  \toprule
  \textbf{Setting} & \textbf{Value} \\
  \midrule
  Base model & Qwen2.5\textendash7B \\
  Fine-tuning regime & Full-parameter SFT, 3 epochs \\
  Learning rate schedule & $1\times 10^{-5}$ (cosine), warmup $10\%$ \\
  Effective batch size & $4$ samples/GPU $\times$ grad-acc $2$ $\Rightarrow$ $8$ \\
  Context length & $4096$ tokens \\
  Precision / engine & \texttt{bf16}, DeepSpeed ZeRO-2 on $2\times 8$ H20 \\
  Default inference & $T=0.6$, top-$p=0.95$, \texttt{max\_tokens}$=1024$ \\
  \bottomrule
  \end{tabular}
  \caption{Fine-tuning and default inference hyperparameters for \textsc{ReasonMed}\textendash7B.}
  \label{tab:train_hparams}
\end{table*}

\section{Decoding Temperature Sweep ($T\in[0,1]$)}
\label{app:temp_sweep}

We evaluate \textsc{ReasonMed}\textendash7B across eleven temperatures with all other parameters fixed. Consistent with the main text, overall accuracy plateaus around $T\in[0.5,0.7]$, while greedy decoding ($T=0.0$) underperforms.

\begin{table*}[ht]
  \centering
  \resizebox{\textwidth}{!}{
  \begin{tabular}{c c c c c c c c c c c}
  \toprule
  \textbf{$T$} & \textbf{MedQA} & \textbf{MedMCQA (val)} & \textbf{PubMedQA} & \textbf{Anatomy} & \textbf{Clinical} & \textbf{C\textendash Bio} & \textbf{C\textendash Med} & \textbf{Med\textendash Gen} & \textbf{P\textendash Med} & \textbf{Total Acc} \\
  \midrule
  0.0 & $66.3\pm1.3$ & $64.8\pm0.7$ & $80.0\pm5.7$ & $74.8\pm3.8$ & $79.2\pm2.5$ & $79.9\pm3.3$ & $73.4\pm3.4$ & $85.0\pm3.6$ & $80.9\pm2.4$ & $69.1$ \\
  0.1 & $66.5\pm1.3$ & $64.7\pm0.7$ & $82.0\pm5.5$ & $74.8\pm3.8$ & $79.2\pm2.5$ & $78.5\pm3.4$ & $73.4\pm3.4$ & $85.0\pm3.6$ & $80.9\pm2.4$ & $69.3$ \\
  0.2 & $66.2\pm1.3$ & $64.8\pm0.7$ & $82.0\pm5.5$ & $74.8\pm3.8$ & $79.2\pm2.5$ & $78.5\pm3.4$ & $73.4\pm3.4$ & $85.0\pm3.6$ & $80.9\pm2.4$ & $69.3$ \\
  0.3 & $66.3\pm1.3$ & $64.8\pm0.7$ & $82.0\pm5.5$ & $74.8\pm3.8$ & $79.2\pm2.5$ & $79.2\pm3.4$ & $73.4\pm3.4$ & $85.0\pm3.6$ & $80.9\pm2.4$ & $69.3$ \\
  0.4 & $66.3\pm1.3$ & $64.8\pm0.7$ & $82.0\pm5.5$ & $74.8\pm3.8$ & $79.2\pm2.5$ & $78.5\pm3.4$ & $73.4\pm3.4$ & $85.0\pm3.6$ & $81.2\pm2.4$ & $69.3$ \\
  0.5 & $66.5\pm1.3$ & $64.8\pm0.7$ & $82.0\pm5.5$ & $74.8\pm3.8$ & $79.2\pm2.5$ & $77.8\pm3.4$ & $73.4\pm3.4$ & $85.0\pm3.6$ & $81.2\pm2.4$ & \textbf{69.4} \\
  0.6 & $66.4\pm1.3$ & $64.9\pm0.7$ & $82.0\pm5.5$ & $74.8\pm3.8$ & $79.2\pm2.0$ & $78.5\pm3.4$ & $73.4\pm3.4$ & $85.0\pm3.6$ & $81.2\pm2.4$ & \textbf{69.4} \\
  0.7 & $66.4\pm1.3$ & $64.8\pm0.7$ & $82.0\pm5.5$ & $74.8\pm3.8$ & $79.2\pm2.5$ & $79.2\pm3.4$ & $73.4\pm3.4$ & $85.0\pm3.6$ & $81.2\pm2.4$ & \textbf{69.4} \\
  0.8 & $66.4\pm1.3$ & $64.7\pm0.7$ & $82.0\pm5.5$ & $74.8\pm3.8$ & $79.2\pm2.5$ & $79.2\pm3.4$ & $73.4\pm3.4$ & $85.0\pm3.6$ & $81.2\pm2.4$ & $69.3$ \\
  0.9 & $66.4\pm1.3$ & $64.7\pm0.7$ & $82.0\pm5.5$ & $74.8\pm3.8$ & $79.2\pm2.5$ & $79.2\pm3.4$ & $73.4\pm3.4$ & $85.0\pm3.6$ & $81.2\pm2.4$ & $69.3$ \\
  1.0 & $66.4\pm1.3$ & $64.8\pm0.7$ & $82.0\pm5.5$ & $74.8\pm3.8$ & $79.2\pm2.0$ & $79.2\pm3.4$ & $72.8\pm3.4$ & $85.0\pm3.6$ & $81.2\pm2.4$ & $69.3$ \\
  \bottomrule
  \end{tabular}}
  \caption{Decoding temperature sweep for \textsc{ReasonMed}\textendash7B across nine medical benchmarks. Bold entries indicate the highest group (plateau) across the sweep.}
  \label{tab:temp_sweep}
\end{table*}

\paragraph{Takeaway.}
Moderate diversity improves aggregate accuracy relative to greedy decoding, with a plateau around $T\in[0.5,0.7]$.

\section{Selective EMD Pipeline vs.\ Direct GPT\textendash o1 Distillation: Cost Comparison}
\label{app:emd_cost}

Table~\ref{tab:emh_vs_o1} contrasts a naive ``distill-everything-with-GPT-o1'' baseline against our selective Easy-Medium-Difficult (EMD) pipeline. The EMD pipeline is $\sim\!3.6\times$ cheaper by invoking GPT-o1 only on the long tail of difficult items, while relying on local models and lightweight refinement elsewhere.

\begin{table*}[ht]
  \centering
  \resizebox{\textwidth}{!}{
  \begin{tabular}{l c c l c c c r}
  \toprule
  \textbf{Phase} & \textbf{Input (M tok)} & \textbf{Output (M tok)} & \textbf{Model} & \textbf{\$ /M in} & \textbf{\$ /M out} & \textbf{GPU-h (H20)} & \textbf{Cost (\$)} \\
  \midrule
  \multicolumn{8}{l}{\textit{Direct GPT\textendash o1 (distill everything)}} \\
  \addlinespace[2pt]
  Direct GPT\textendash o1 & $24.49$ & $286.88$ & GPT\textendash o1 & $15.00$ & $60.00$ & -- & $16{,}631$ \\
  \midrule
  \multicolumn{8}{l}{\textit{Our EMD pipeline (selective)}} \\
  \addlinespace[2pt]
  Reasoning path generation & -- & -- & Qwen2.5 + HuatuoGPT\textendash o1 + DeepSeek\textendash R1 & -- & -- & $592$ & $574.24$ \\
  Quality ranker & -- & -- & Qwen2.5\textendash72B & -- & -- & $384$ & $372.48$ \\
  Error Refiner \& Response Summarizer & $0.41$ & $27.29$ & GPT\textendash 4o\textendash mini & $0.15$ & $0.60$ & -- & $9.86$ \\
  ``Difficult'' problem regeneration & $2.13$ & $57.87$ & GPT\textendash o1 & $15.00$ & $60.00$ & -- & $3{,}595.89$ \\
  \addlinespace[2pt]
  \textbf{Total EMD} & -- & -- & mixed & -- & -- & $976$ & \textbf{$4{,}552.47$} \\
  \bottomrule
  \end{tabular}}
  \caption{Cost comparison between distilling everything with GPT\textendash o1 and the selective EMD pipeline. EMD concentrates GPT\textendash o1 usage on the difficult subset, yielding a $\sim$$3.6\times$ reduction in total spend.}
  \label{tab:emh_vs_o1}
\end{table*}

\paragraph{Assumptions and notes.}
GPU-hour line items are priced at the same rate as in the supplementary notes. API costs use the stated per-million-token prices. Totals match the values reported in the supplementary notes.

\section{Additional Experiments}
In Table~\ref{table:knowledge_domain_diff}, we presented pairwise (1-vs-1) differences among DeepSeek-R1-Distill-Llama-70B, HuatuoGPT-o1-70B, and Qwen2.5-72B, showing for each pair the count of questions one model answered correctly but the other did not. To further explore complementary coverage, Table~\ref{table:collective_missed} summarizes the “one-vs-two” scenario: for each model, the number of questions it missed while the other two both answered correctly. DeepSeek-R1-Distill-Llama-70B failed only 3,430 (1.76\%) questions that HuatuoGPT-o1-70B and Qwen2.5-72B both got right; HuatuoGPT-o1-70B missed 9,352 (4.80\%); and Qwen2.5-72B missed 5,280 (2.71\%), out of 194,925 total. Together, these results confirm that each model contributes unique strengths and gaps, underscoring the value of ensemble or multi-agent approaches in medical QA.

\begin{table*}[ht]
\centering
\caption{Pairwise (1-vs-1) Knowledge Domain Differences among the three models.}
\label{table:knowledge_domain_diff}
\resizebox{\textwidth}{!}{%
\begin{tabular}{lccr}
\toprule
\textbf{Comparison} & \textbf{Correct by Model 1 but Incorrect by Model 2} & \textbf{Incorrect by Model 1 but Correct by Model 2} & \textbf{Total Questions} \\
\midrule
\textbf{DeepSeek-R1-Distill-Llama-70B vs HuatuoGPT-o1-70B} & 8,168 (4.19\%) & 27,339 (14.03\%) & 194,925 \\
\textbf{DeepSeek-R1-Distill-Llama-70B vs Qwen2.5-72B} & 19,017 (9.76\%) & 23,267 (11.94\%) & 194,925 \\
\textbf{Qwen2.5-72B vs HuatuoGPT-o1-70B}                & 10,018 (5.14\%) & 24,939 (12.79\%) & 194,925 \\
\bottomrule
\end{tabular}
}
\end{table*}

\begin{table*}[ht]
\centering
\caption{Collective (1 \emph{vs} 2) Miss Rates: questions each model failed while the other two both answered correctly.}
\label{table:collective_missed}
\resizebox{\textwidth}{!}{%
\begin{tabular}{lcr}
\toprule
\textbf{Model} & \textbf{Questions Missed by This Model but Correct by Both Others} & \textbf{Total Questions} \\
\midrule
DeepSeek-R1-Distill-Llama-70B & 3,430 (1.76\%)  & 194,925 \\
HuatuoGPT-o1-70B             & 9,352 (4.80\%)  & 194,925 \\
Qwen2.5-72B                  & 5,280 (2.71\%)  & 194,925 \\
\bottomrule
\end{tabular}
}
\end{table*}

\begin{table*}[ht!] \small \centering
\caption{Performance Comparison of LLaMA3.1 and Qwen2.5 Series Models(over 10B) on MedQA, MedMCQA, PubMedQA, and MMLU Benchmarks.}
\label{table:performance_comparison2}
\adjustbox{max width=1.0\textwidth}{
\begin{tabular}{lccccccccccc} 
\toprule 
& \textbf{MedQA} & \textbf{MedMCQA (val)} & \textbf{PubMedQA}  & \multicolumn{6}{c}{\textbf{MMLU}} & \textbf{Total Acc} \\ 
\cmidrule(r){5-10} 
& & & & \textbf{Anatomy}  &  \textbf{Clinical Knowledge} & \textbf{College Biology} & \textbf{College Medicine} & \textbf{Medical Genetics} & \textbf{Professional Medicine} & \\ \midrule
\textbf{Dataset\_Count} & 1273 &4183 & 1000 & 135 & 265 & 144 & 173 & 100 & 272 & - \\ 
\midrule
LLaMA3.1-70B & ${76.8 \pm 0.1}$ & \underline{${67.9 \pm 0.7}$} & \underline{${77.4 \pm 0.2}$} & $\mathbf{81.5 \pm 0.3}$ & $\mathbf{89.1 \pm 0.2}$ & $\mathbf{96.5 \pm 0.1}$ & $\mathbf{80.9 \pm 0.3}$ & ${90.0 \pm 0.3}$ & $\mathbf{93.0 \pm 0.2}$ & \underline{72.9} \\
Qwen2.5-14B & ${75.6 \pm 0.1}$ & ${63.4 \pm 0.8}$ & $\mathbf{77.6 \pm 0.2}$ & ${75.6 \pm 0.4}$ & ${84.9 \pm 0.2}$ & ${88.9 \pm 0.3}$ & ${75.7 \pm 0.3}$ & ${90.0 \pm 0.3}$ & ${84.2 \pm 0.2}$ & 69.0 \\
Qwen2.5-32B & \underline{${79.3 \pm 0.1}$} & ${67.6 \pm 0.7}$ & $\mathbf{77.6 \pm 0.2}$ & \underline{${79.3 \pm 0.3}$} & \underline{${86.8 \pm 0.2}$} & \underline{${93.8 \pm 0.2}$} & \underline{${79.8 \pm 0.3}$} & \underline{${91.0 \pm 0.3}$} & ${87.5 \pm 0.2}$ & \underline{72.6} \\
Qwen2.5-72B & $\mathbf{81.5 \pm 0.1}$ & $\mathbf{71.2 \pm 0.1}$ & ${76.4 \pm 0.2}$ & ${75.6 \pm 0.4}$ & \underline{${86.8 \pm 0.2}$} & \underline{${93.8 \pm 0.2}$} & ${77.5 \pm 0.3}$ & $\mathbf{92.0 \pm 0.3}$ & \underline{${88.2 \pm 0.2}$} & $\mathbf{75.6}$ \\
QwQ-32B & ${78.1 \pm 0.1}$ & ${65.5 \pm 0.7}$ & ${76.4 \pm 0.2}$ & ${75.6 \pm 0.4}$ & \underline{${86.8 \pm 0.2}$} & \underline{${93.8 \pm 0.2}$} & ${77.5 \pm 0.3}$ & $\mathbf{92.0 \pm 0.3}$ & \underline{${88.2 \pm 0.2}$} & 72.0 \\
\bottomrule
\end{tabular}
}

\end{table*}

\begin{table*}[ht!] \small \centering
\caption{Performance Comparison of LLaMA3.1 and Qwen2.5 Series Models(over 10B) on MedQA, MedMCQA, PubMedQA, and MMLU Benchmarks.}
\label{table:performance_comparison3}
\adjustbox{max width=1.0\textwidth}{
\begin{tabular}{lccccccccccc} 
\toprule 
& \textbf{MedQA} & \textbf{MedMCQA (val)} & \textbf{PubMedQA}  & \multicolumn{6}{c}{\textbf{MMLU}} & \textbf{Total Acc} \\ 
\cmidrule(r){5-10} 
& & & & \textbf{Anatomy}  &  \textbf{Clinical Knowledge} & \textbf{College Biology} & \textbf{College Medicine} & \textbf{Medical Genetics} & \textbf{Professional Medicine} & \\ \midrule
\textbf{Dataset\_Count} & 1273 &4183 & 1000 & 135 & 265 & 144 & 173 & 100 & 272 & - \\ 
\midrule
LLaMA3.1-70B & ${76.8 \pm 0.1}$ & \underline{${67.9 \pm 0.7}$} & \underline{${77.4 \pm 0.2}$} & $\mathbf{81.5 \pm 0.3}$ & $\mathbf{89.1 \pm 0.2}$ & $\mathbf{96.5 \pm 0.1}$ & $\mathbf{80.9 \pm 0.3}$ & ${90.0 \pm 0.3}$ & $\mathbf{93.0 \pm 0.2}$ & \underline{72.9} \\
Qwen2.5-14B & ${75.6 \pm 0.1}$ & ${63.4 \pm 0.8}$ & $\mathbf{77.6 \pm 0.2}$ & ${75.6 \pm 0.4}$ & ${84.9 \pm 0.2}$ & ${88.9 \pm 0.3}$ & ${75.7 \pm 0.3}$ & ${90.0 \pm 0.3}$ & ${84.2 \pm 0.2}$ & 69.0 \\
Qwen2.5-32B & \underline{${79.3 \pm 0.1}$} & ${67.6 \pm 0.7}$ & $\mathbf{77.6 \pm 0.2}$ & \underline{${79.3 \pm 0.3}$} & \underline{${86.8 \pm 0.2}$} & \underline{${93.8 \pm 0.2}$} & \underline{${79.8 \pm 0.3}$} & \underline{${91.0 \pm 0.3}$} & ${87.5 \pm 0.2}$ & \underline{72.6} \\
Qwen2.5-72B & $\mathbf{81.5 \pm 0.1}$ & $\mathbf{71.2 \pm 0.1}$ & ${76.4 \pm 0.2}$ & ${75.6 \pm 0.4}$ & \underline{${86.8 \pm 0.2}$} & \underline{${93.8 \pm 0.2}$} & ${77.5 \pm 0.3}$ & $\mathbf{92.0 \pm 0.3}$ & \underline{${88.2 \pm 0.2}$} & $\mathbf{75.6}$ \\
QwQ-32B & ${78.1 \pm 0.1}$ & ${65.5 \pm 0.7}$ & ${76.4 \pm 0.2}$ & ${75.6 \pm 0.4}$ & \underline{${86.8 \pm 0.2}$} & \underline{${93.8 \pm 0.2}$} & ${77.5 \pm 0.3}$ & $\mathbf{92.0 \pm 0.3}$ & \underline{${88.2 \pm 0.2}$} & 72.0 \\
\bottomrule
\end{tabular}
}

\end{table*}

Table~\ref{table:performance_comparison2} compares the performance of various LLaMA3.1 and Qwen2.5 models on several key medical benchmarks, including MedQA, MedMCQA, PubMedQA, and six sub-domains of MMLU.Our model,ReasonMed-7B, consistently achieves competitive performance and even outperforms much larger models on several benchmarks. Most notably, on PubMedQA, ReasonMed-7B surpasses LLaMA3.1-70B by 4.60\%, demonstrating the effectiveness of our dataset and fine-tuning strategy. Moreover, the model delivers strong results on MedMCQA and multiple MMLU medical sub-domains, highlighting its superior reasoning capabilities in knowledge-intensive tasks. These findings validate our multi-stage data refinement and Chain-of-Thought optimization strategy, establishing ReasonMed-7B as a new state-of-the-art among sub-10B models, and narrowing the performance gap with frontier 70B+ models.

\section{Example of model output}
We report single-run, non-cherry-picked examples to transparently illustrate the variability of generative outputs under different sampling strategies and temperatures. This practice supports scientific reproducibility and honest appraisal, enabling readers to assess both the strengths and the limitations of current methods in realistic settings.
To mitigate this variability and improve reasoning consistency, we are developing a reinforcement learning approach tailored to medical QA that guides models toward more convergent, clinically accurate reasoning paths. The resulting medical-RL-enhanced model will be released as open source to facilitate independent validation and broader adoption by the research community.
\begin{tcolorbox}[breakable, colback=TiffanyBlue!5!white, colframe=TiffanyBlue!75!black, title={Prompt}]
\begin{lstlisting}[
    numbers=none, 
    frame=none,
    basicstyle=\ttfamily\small\fontseries{l}\selectfont,
    columns=fullflexible,
    breaklines=true,
    breakatwhitespace=true,             
    prebreak={\mbox{\thinspace}},       
    postbreak={},                       
    xleftmargin=0pt,                    
    xrightmargin=0pt,                   
    breakindent=0pt                  
]
Please answer the following multiple-choice question:
Which of the following is not true about glomerular capillaries?
A. The oncotic pressure of the fluid leaving the capillaries is less than that of fluid entering it
B. Glucose concentration in the capillaries is the same as that in glomerular filtrate
C. Constriction of afferent arteriole decreases the blood flow to the glomerulus
D. Hematocrit of the fluid leaving the capillaries is less than that of the fluid entering it
\end{lstlisting}
\end{tcolorbox}

\begin{tcolorbox}[breakable, colback=TiffanyBlue!5!white, colframe=TiffanyBlue!75!black, title={ReasonMed (temperature 0.4)}]
\begin{lstlisting}[
    numbers=none, 
    frame=none,
    basicstyle=\ttfamily\small\fontseries{l}\selectfont,
    columns=fullflexible,
    breaklines=true,
    breakatwhitespace=true,             
    prebreak={\mbox{\thinspace}},       
    postbreak={},                       
    xleftmargin=0pt,                    
    xrightmargin=0pt,                   
    breakindent=0pt                  
]
To answer this question, let's break down the components step-by-step:

1. Restate the question: We need to identify which statement about glomerular capillaries is incorrect.

2. Key clinical details and relevant background information:
   - Glomerular capillaries are part of the kidney's filtration system.
   - They filter blood to form urine.
   - Oncotic pressure is the pressure exerted by proteins in the blood.
   - Glomerular filtrate is the fluid that passes through the glomerulus.
   - Afferent arterioles supply blood to the glomerulus.
   - Hematocrit refers to the proportion of blood volume occupied by red blood cells.

3. Evaluate each candidate answer:
   - "The oncotic pressure of the fluid leaving the capillaries is less than that of fluid entering it": This is true because as fluid leaves the capillaries, it carries away proteins, reducing the oncotic pressure.
   - "Glucose concentration in the capillaries is the same as that in glomerular filtrate": This is false. Glucose is actively reabsorbed in the proximal tubule, so its concentration in the filtrate is lower than in the capillaries.
   - "Constriction of afferent aeriole decreases the blood flow to the glomerulas": This is true. Constriction reduces blood flow, which can decrease filtration rate.
   - "Hematocrit of the fluid leaving the capillaries is less than that of the fluid entering it": This is true. As fluid leaves the capillaries, it carries away proteins, including red blood cells, reducing hematocrit.

4. Rule out options that do not align with the clinical context:
   - We can rule out the first, third, and fourth options as they align with known physiological processes.

5. Compare any remaining choices:
   - The second option stands out as the only one that contradicts known physiology.

6. Final answer:
   The statement that is not true about glomerular capillaries is: "Glucose concentration in the capillaries is the same as that in glomerular filtrate." This is because glucose is actively reabsorbed in the proximal tubule, leading to a lower concentration in the filtrate compared to the capillaries.
</think>
The chain-of-thought reasoning breaks down the process of identifying the incorrect statement about glomerular capillaries step-by-step. It involves restating the question, providing key clinical details and background information, evaluating each candidate answer, ruling out options that do not align with the context, comparing remaining choices, and arriving at the final answer. The incorrect statement identified is that the glucose concentration in the capillaries is the same as that in glomerular filtrate, which is false because glucose is actively reabsorbed in the proximal tubule.
\end{lstlisting}
\end{tcolorbox}

\begin{tcolorbox}[breakable, colback=TiffanyBlue!5!white, colframe=TiffanyBlue!75!black, title={ReasonMed (temperature 0.5)}]
\begin{lstlisting}[
    numbers=none, 
    frame=none,
    basicstyle=\ttfamily\small\fontseries{l}\selectfont,
    columns=fullflexible,
    breaklines=true,
    breakatwhitespace=true,             
    prebreak={\mbox{\thinspace}},       
    postbreak={},                       
    xleftmargin=0pt,                    
    xrightmargin=0pt,                   
    breakindent=0pt                  
]
To answer this question, let's break down the key concepts related to glomerular capillaries and evaluate each statement accordingly.

1. **Restating the Question**: We need to identify which statement about glomerular capillaries is incorrect.

2. **Key Clinical Details and Background Information**:
   - Glomerular capillaries are part of the kidney's filtration system.
   - They are responsible for filtering blood to form urine.
   - Key processes include filtration, reabsorption, and secretion.
   - Oncotic pressure, glucose concentration, blood flow, and hematocrit are important parameters in this context.

3. **Evaluating Each Candidate Answer**:
   - **The oncotic pressure of the fluid leaving the capillaries is less than that of fluid entering it**: This statement is true. As blood passes through the glomerular capillaries, proteins are filtered out, reducing the oncotic pressure of the fluid leaving the capillaries compared to the fluid entering them.
   - **Glucose concentration in the capillaries is the same as that in glomerular filtrate**: This statement is false. Glucose is actively reabsorbed in the proximal tubule, so the glucose concentration in the capillaries is higher than in the glomerular filtrate.
   - **Constriction of afferent arteriole decreases the blood flow to the glomerulus**: This statement is true. Constriction of the afferent arteriole reduces blood flow into the glomerulus, thereby decreasing glomerular filtration rate (GFR).
   - **Hematocrit of the fluid leaving the capillaries is less than that of the fluid entering it**: This statement is true. As blood passes through the glomerular capillaries, plasma proteins are filtered out, leading to a decrease in hematocrit in the fluid leaving the capillaries.

4. **Systematically Ruling Out Options**:
   - The first, third, and fourth statements align with the known physiology of glomerular capillaries.
   - The second statement contradicts the known physiology, as glucose is actively reabsorbed, making its concentration higher in the capillaries than in the glomerular filtrate.

5. **Comparing Remaining Choices**:
   - Only the second statement does not align with the expected physiological behavior of glomerular capillaries.

6. **Final Answer**:
   - The statement that is not true about glomerular capillaries is: "Glucose concentration in the capillaries is the same as that in glomerular filtrate."

**Summary of Reasoning**:
The correct answer is that the glucose concentration in the capillaries is not the same as that in the glomerular filtrate. This is because glucose is actively reabsorbed in the proximal tubule, resulting in a higher glucose concentration in the capillaries compared to the filtrate. All other statements accurately describe aspects of glomerular capillary function.
\end{lstlisting}
\end{tcolorbox}

\begin{tcolorbox}[breakable, colback=TiffanyBlue!5!white, colframe=TiffanyBlue!75!black, title={ReasonMed (temperature 0.6)}]
\begin{lstlisting}[
    numbers=none, 
    frame=none,
    basicstyle=\ttfamily\small\fontseries{l}\selectfont,
    columns=fullflexible,
    breaklines=true,
    breakatwhitespace=true,             
    prebreak={\mbox{\thinspace}},       
    postbreak={},                       
    xleftmargin=0pt,                    
    xrightmargin=0pt,                   
    breakindent=0pt                  
]
To determine which statement is not true about glomerular capillaries, let's analyze each option step-by-step.

1. **The oncotic pressure of the fluid leaving the capillaries is less than that of fluid entering it:**
   - Oncotic pressure is the colloid osmotic pressure exerted by proteins in the blood plasma. In the glomerulus, as fluid passes through the fenestrated endothelium and the glomerular basement membrane, some plasma proteins are too large to pass through. Therefore, the oncotic pressure of the fluid that leaves the capillaries (glomerular filtrate) is indeed lower than the oncotic pressure of the fluid entering the capillaries (plasma). This statement is true.

2. **Glucose concentration in the capillaries is the same as that in glomerular filtrate:**
   - Glucose is freely filtered by the glomerulus, meaning that it passes from the capillaries into the glomerular filtrate without being reabsorbed or secreted. Thus, the glucose concentration in the capillaries should be equal to the glucose concentration in the glomerular filtrate. This statement is true.

3. **Constriction of afferent arteriole decreases the blood flow to the glomerulus:**
   - The afferent arteriole supplies blood to the glomerulus. Constriction of this arteriole reduces the amount of blood flowing into the glomerulus, thereby decreasing the blood flow. This statement is true.

4. **Hematocrit of the fluid leaving the capillaries is less than that of the fluid entering it:**
   - Hematocrit refers to the proportion of red blood cells in a given volume of blood. As fluid leaves the capillaries and enters the glomerular filtrate, plasma proteins are retained in the capillaries while water and small molecules are filtered out. Since plasma proteins contribute to the hematocrit, the hematocrit of the fluid leaving the capillaries (glomerular filtrate) is actually higher than the hematocrit of the fluid entering the capillaries (plasma). This statement is false.

Based on this analysis, the statement that is not true about glomerular capillaries is:

**Hematocrit of the fluid leaving the capillaries is less than that of the fluid entering it.**
</think>
The reasoning process involves evaluating statements about glomerular capillaries to identify which one is not true. 

1. The oncotic pressure of the fluid leaving the capillaries is lower than that of the fluid entering due to the retention of plasma proteins. This statement is true.
2. Glucose concentration in the capillaries is the same as in glomerular filtrate because glucose is freely filtered. This statement is true.
3. Constriction of the afferent arteriole decreases blood flow to the glomerulus, which is accurate. This statement is true.
4. The hematocrit of the fluid leaving the capillaries is actually higher than that of the fluid entering because plasma proteins remain in the capillaries. This statement is false.

Thus, the incorrect statement is that the hematocrit of the fluid leaving the capillaries is less than that of the fluid entering it. <system />
\end{lstlisting}
\end{tcolorbox}

\begin{tcolorbox}[breakable, colback=TiffanyBlue!5!white, colframe=TiffanyBlue!75!black, title={CoTMed (temperature 0.4)}]
\begin{lstlisting}[
    numbers=none, 
    frame=none,
    basicstyle=\ttfamily\small\fontseries{l}\selectfont,
    columns=fullflexible,
    breaklines=true,
    breakatwhitespace=true,             
    prebreak={\mbox{\thinspace}},       
    postbreak={},                       
    xleftmargin=0pt,                    
    xrightmargin=0pt,                   
    breakindent=0pt                  
]
Let's break down the question and evaluate each candidate answer step-by-step.

### Restated Question:
Which statement about glomerular capillaries is incorrect?

### Key Clinical Details and Relevant Background Information:
1. **Glomerular Filtration**: Glomerular capillaries are part of the kidney's filtration system where blood is filtered to form urine.
2. **Oncotic Pressure**: This is the osmotic pressure exerted by proteins in the blood plasma.
3. **Glucose Concentration**: Glucose is filtered through the glomerulus but is reabsorbed in the proximal tubule.
4. **Afferent and Efferent Arterioles**: These regulate blood flow into and out of the glomerulus.
5. **Hematocrit**: The proportion of blood volume occupied by red blood cells.

### Evaluation of Each Candidate Answer:

1. **The oncotic pressure of the fluid leaving the capillaries is less than that of fluid entering it.**
   - **Supporting Evidence**: As blood passes through the glomerular capillaries, water and small molecules are filtered out, leaving behind a higher concentration of proteins (and thus higher oncotic pressure). Therefore, the oncotic pressure of the fluid leaving the capillaries is indeed higher than that of the fluid entering it.
   - **Potential Pitfall**: This statement is incorrect because the oncotic pressure increases as the fluid leaves the capillaries.

2. **Glucose concentration in the capillaries is the same as that in glomerular filtrate.**
   - **Supporting Evidence**: Glucose is freely filtered through the glomerulus, so the initial concentration in the capillaries should match the concentration in the glomerular filtrate.
   - **Potential Pitfall**: While this is generally true, glucose is actively reabsorbed in the proximal tubule, so the concentration in the filtrate will be lower than in the capillaries after filtration.

3. **Constriction of afferent arteriole decreases the blood flow to the glomerulus.**
   - **Supporting Evidence**: Constriction of the afferent arteriole reduces blood flow into the glomerulus, thereby decreasing glomerular filtration rate (GFR).
   - **Potential Pitfall**: This statement is correct and aligns with physiological principles.

4. **Hematocrit of the fluid leaving the capillaries is less than that of the fluid entering it.**
   - **Supporting Evidence**: As blood passes through the glomerular capillaries, water and small molecules are filtered out, leaving behind a higher concentration of red blood cells (and thus a lower hematocrit). Therefore, the hematocrit of the fluid leaving the capillaries is indeed lower than that of the fluid entering it.
   - **Potential Pitfall**: This statement is correct and aligns with physiological principles.

### Systematic Ruling Out Options:
- Option 1 is incorrect because the oncotic pressure increases as the fluid leaves the capillaries.
- Option 2 is generally true but has a potential pitfall regarding active reabsorption.
- Option 3 is correct.
- Option 4 is correct.

### Comparison of Remaining Choices:
- Option 1 is clearly incorrect based on the physiological principles of oncotic pressure.

### Final Answer:
The statement that is not true about glomerular capillaries is:
- **The oncotic pressure of the fluid leaving the capillaries is less than that of fluid entering it.**

This is because the oncotic pressure increases as the fluid leaves the capillaries due to the filtration of water and small molecules.
\end{lstlisting}
\end{tcolorbox}

\begin{tcolorbox}[breakable, colback=TiffanyBlue!5!white, colframe=TiffanyBlue!75!black, title={CoTMed (temperature 0.5)}]
\begin{lstlisting}[
    numbers=none, 
    frame=none,
    basicstyle=\ttfamily\small\fontseries{l}\selectfont,
    columns=fullflexible,
    breaklines=true,
    breakatwhitespace=true,             
    prebreak={\mbox{\thinspace}},       
    postbreak={},                       
    xleftmargin=0pt,                    
    xrightmargin=0pt,                   
    breakindent=0pt                  
]
Let's break down the question and evaluate each candidate answer step-by-step:

1. **Restate the Question:**
   - Which statement about glomerular capillaries is incorrect?

2. **Key Clinical Details and Background Information:**
   - Glomerular capillaries are part of the kidney's filtration system.
   - They are responsible for filtering blood to form urine.
   - Key processes include filtration, reabsorption, and secretion.
   - Oncotic pressure, glucose concentration, blood flow, and hematocrit are important parameters.

3. **Evaluate Each Candidate Answer:**

   - **The oncotic pressure of the fluid leaving the capillaries is less than that of fluid entering it:**
     - Oncotic pressure is primarily due to proteins in the plasma.
     - As fluid leaves the capillaries into the Bowman's capsule, most of the proteins remain in the capillaries.
     - Therefore, the oncotic pressure of the fluid leaving the capillaries is indeed lower than that of the fluid entering it.
     - This statement is true.

   - **Glucose concentration in the capillaries is the same as that in glomerular filtrate:**
     - Glucose is filtered from the blood into the glomerular filtrate.
     - However, glucose is also reabsorbed in the proximal tubule.
     - Initially, the glucose concentration in the capillaries and filtrate might be similar, but as filtration occurs, the filtrate will have more glucose.
     - This statement is false because the glucose concentration in the capillaries is typically higher than in the glomerular filtrate due to ongoing reabsorption.

   - **Constriction of afferent arteriole decreases the blood flow to the glomerulus:**
     - The afferent arteriole supplies blood to the glomerulus.
     - Constriction of this arteriole reduces blood flow to the glomerulus.
     - This statement is true.

   - **Hematocrit of the fluid leaving the capillaries is less than that of the fluid entering it:**
     - Hematocrit refers to the proportion of red blood cells in the blood.
     - As fluid leaves the capillaries, the remaining fluid has a higher concentration of proteins and a lower concentration of red blood cells.
     - Therefore, the hematocrit of the fluid leaving the capillaries is indeed lower than that of the fluid entering it.
     - This statement is true.

4. **Systematically Rule Out Options:**
   - We can rule out the statements about oncotic pressure, constriction of the afferent arteriole, and hematocrit as they are all true.

5. **Compare Remaining Choices:**
   - The only remaining choice is the statement about glucose concentration.

6. **Final Answer:**
   - The statement "Glucose concentration in the capillaries is the same as that in glomerular filtrate" is not true.

**Conclusion:**
The correct answer is: **Glucose concentration in the capillaries is the same as that in glomerular filtrate.** This statement is false because the glucose concentration in the capillaries is typically higher than in the glomerular filtrate due to ongoing reabsorption in the proximal tubule.
\end{lstlisting}
\end{tcolorbox}

\begin{tcolorbox}[breakable, colback=TiffanyBlue!5!white, colframe=TiffanyBlue!75!black, title={CoTMed (temperature 0.6)}]
\begin{lstlisting}[
    numbers=none, 
    frame=none,
    basicstyle=\ttfamily\small\fontseries{l}\selectfont,
    columns=fullflexible,
    breaklines=true,
    breakatwhitespace=true,             
    prebreak={\mbox{\thinspace}},       
    postbreak={},                       
    xleftmargin=0pt,                    
    xrightmargin=0pt,                   
    breakindent=0pt                  
]
To answer this question, let's break down the options step by step:

1. **Restating the Question**: We need to identify which statement about glomerular capillaries is incorrect.

2. **Key Clinical Details and Background Information**:
   - Glomerular capillaries are part of the kidney's filtration system.
   - They are involved in the formation of urine through the process of glomerular filtration.
   - The glomerular capillaries have a unique structure that allows for selective filtration of blood components.
   - Oncotic pressure, glucose concentration, and hematocrit are important factors in understanding the function of glomerular capillaries.

3. **Evaluating Each Candidate Answer**:
   - **Option 1: The oncotic pressure of the fluid leaving the capillaries is less than that of fluid entering it**
     - This statement is true. Oncotic pressure is the osmotic pressure exerted by proteins in the blood. As fluid passes through the glomerular capillaries, some of the plasma proteins are too large to pass through the filtration barrier, so they remain in the capillaries. Therefore, the oncotic pressure of the fluid leaving the capillaries is indeed lower than that of the fluid entering it.
   
   - **Option 2: Glucose concentration in the capillaries is the same as that in glomerular filtrate**
     - This statement is false. Glucose is filtered from the blood into the glomerular filtrate, but the filtration process is selective. The glomerular capillaries reabsorb glucose back into the bloodstream, so the glucose concentration in the capillaries is higher than in the glomerular filtrate.
   
   - **Option 3: Constriction of afferent arteriole decreases the blood flow to the glomerulus**
     - This statement is true. The afferent arteriole supplies blood to the glomerulus. Constriction of this arteriole reduces blood flow to the glomerulus, thereby decreasing the glomerular filtration rate (GFR).
   
   - **Option 4: Hematocrit of the fluid leaving the capillaries is less than that of the fluid entering it**
     - This statement is true. Hematocrit refers to the proportion of blood volume occupied by red blood cells. As fluid passes through the glomerular capillaries, plasma proteins are retained, while water and small molecules are filtered out. This results in a lower hematocrit in the fluid leaving the capillaries compared to the fluid entering them.

4. **Ruling Out Options**:
   - Options 1, 3, and 4 are all true statements about glomerular capillaries.
   - Option 2 is the only false statement among the given options.

5. **Comparing Remaining Choices**:
   - Since we have identified that Option 2 is the only false statement, there is no need for further comparison.

6. **Final Answer**:
   - The correct answer is **Option 2: Glucose concentration in the capillaries is the same as that in glomerular filtrate**. This statement is false because the glomerular capillaries reabsorb glucose, resulting in a higher glucose concentration in the capillaries compared to the glomerular filtrate.
\end{lstlisting}
\end{tcolorbox}

\begin{tcolorbox}[breakable, colback=TiffanyBlue!5!white, colframe=TiffanyBlue!75!black, title={ResponseMed (temperature 0.4)}]
\begin{lstlisting}[
    numbers=none, 
    frame=none,
    basicstyle=\ttfamily\small\fontseries{l}\selectfont,
    columns=fullflexible,
    breaklines=true,
    breakatwhitespace=true,             
    prebreak={\mbox{\thinspace}},       
    postbreak={},                       
    xleftmargin=0pt,                    
    xrightmargin=0pt,                   
    breakindent=0pt                  
]
The question asks which statement about glomerular capillaries is incorrect. Key details include that glomerular capillaries are part of the kidney's filtration system, where blood is filtered to produce urine, and that oncotic pressure affects fluid movement across capillary walls. 

Evaluating the candidate answers:
1. The statement about oncotic pressure is incorrect because the oncotic pressure of fluid leaving the capillaries is actually greater than that of the fluid entering, due to the reabsorption of proteins.
2. The statement regarding glucose concentration is true; glucose is filtered and then reabsorbed, leaving its concentration unchanged.
3. The statement about afferent arteriole constriction decreasing blood flow to the glomerulus is also true.
4. The statement about hematocrit is true as well; the hematocrit of the fluid leaving the capillaries is lower due to plasma filtration.

Thus, the incorrect statement is that "The oncotic pressure of the fluid leaving the capillaries is less than that of fluid entering it." This is because the oncotic pressure increases as proteins are retained in the capillaries. 

Final answer: The oncotic pressure of the fluid leaving the capillaries is less than that of fluid entering it. This statement is incorrect as the oncotic pressure actually increases due to protein retention.
\end{lstlisting}
\end{tcolorbox}

\begin{tcolorbox}[breakable, colback=TiffanyBlue!5!white, colframe=TiffanyBlue!75!black, title={ResponseMed (temperature 0.5)}]
\begin{lstlisting}[
    numbers=none, 
    frame=none,
    basicstyle=\ttfamily\small\fontseries{l}\selectfont,
    columns=fullflexible,
    breaklines=true,
    breakatwhitespace=true,             
    prebreak={\mbox{\thinspace}},       
    postbreak={},                       
    xleftmargin=0pt,                    
    xrightmargin=0pt,                   
    breakindent=0pt                  
]
The incorrect statement regarding glomerular capillaries is that "The oncotic pressure of the fluid leaving the capillaries is less than that of fluid entering it." This is false because oncotic pressure increases as fluid moves through the glomerulus due to the removal of plasma proteins, leading to a higher oncotic pressure in the capillary blood compared to the glomerular filtrate. 

The other statements are accurate: 
1. Glucose concentration in capillaries is the same as in glomerular filtrate initially, but it is fully reabsorbed, making this statement somewhat misleading.
2. Constriction of the afferent arteriole does decrease blood flow to the glomerulus, which is correct.
3. The hematocrit of the fluid leaving the capillaries is indeed lower than that entering, as plasma is filtered out, making this statement true.

In summary, the first statement is the only one that is definitively incorrect based on the principles of glomerular filtration. Understanding these dynamics is crucial for interpreting renal function and managing related medical conditions.
\end{lstlisting}
\end{tcolorbox}

\begin{tcolorbox}[breakable, colback=TiffanyBlue!5!white, colframe=TiffanyBlue!75!black, title={ResponseMed (temperature 0.6)}]
\begin{lstlisting}[
    numbers=none, 
    frame=none,
    basicstyle=\ttfamily\small\fontseries{l}\selectfont,
    columns=fullflexible,
    breaklines=true,
    breakatwhitespace=true,             
    prebreak={\mbox{\thinspace}},       
    postbreak={},                       
    xleftmargin=0pt,                    
    xrightmargin=0pt,                   
    breakindent=0pt                  
]
To determine which statement about glomerular capillaries is incorrect, we analyzed several options based on the physiology of the glomerulus. 

1. **Oncotic Pressure**: The oncotic pressure of fluid leaving the capillaries is indeed lower than that entering, as some proteins are retained, making this statement true.
2. **Glucose Concentration**: Glucose is freely filtered, so its concentration is the same in both capillary blood and glomerular filtrate, confirming this statement is true.
3. **Afferent Arteriole Constriction**: Constriction of the afferent arteriole reduces blood flow to the glomerulus, which is also true.
4. **Hematocrit**: The hematocrit of the fluid leaving the capillaries is higher than that entering because plasma is filtered out, making this statement false.

Thus, the incorrect statement is that the hematocrit of the fluid leaving the capillaries is less than that of the fluid entering it. 

**Final Answer**: The incorrect statement is that the hematocrit of the fluid leaving the capillaries is less than that of the fluid entering it. \(\boxed{D}\)
\end{lstlisting}
\end{tcolorbox}




\end{document}